%% file: arxiv/Verdi_arxiv.tex
\pdfoutput=1
\documentclass[conference
]{IEEEtran}
\input{cvpr/preamble}

\usepackage{cite}
\usepackage{makecell}
\usepackage[utf8]{inputenc}
\usepackage{amsmath, amssymb, amsfonts}
\usepackage{lineno}
\usepackage{algorithmic}
\usepackage{algorithm}
\usepackage{graphicx}
\usepackage{textcomp}
\usepackage[usenames,dvipsnames]{xcolor}
\usepackage{xspace}
\usepackage{booktabs}
\usepackage{multirow}
\usepackage{url}
\usepackage{cite}
\usepackage{bbm}
\usepackage{comment}
\usepackage{subcaption}
\usepackage{mathtools}
\usepackage{balance}
\usepackage{hyperref}
\hypersetup{
    colorlinks=true,
    linkcolor=black,
    citecolor=black,
    urlcolor=blue,
}
\usepackage[nameinlink,noabbrev]{cleveref}
\usepackage{wrapfig}
\usepackage{tcolorbox}
\tcbuselibrary{listingsutf8}
\usepackage{listings}

\usepackage[protrusion=true,
    activate={true,nocompatibility},
    final,
    tracking=true,
    kerning=true,
    spacing=false,
    factor=1100]{microtype}
    
\SetTracking{encoding={*}, shape=sc}{10}
\usepackage{amsthm}

\newcommand\BibTeX{{\rmfamily B\kern-.05em \textsc{i\kern-.025em b}\kern-.08em
T\kern-.1667em\lower.7ex\hbox{E}\kern-.125emX}}

\tcbset{
  promptstyle/.style={
    colback=gray!10,
    colframe=gray!60,
    boxrule=0.5pt,
    arc=2pt,
    outer arc=2pt,
    left=4pt,
    right=4pt,
    top=4pt,
    bottom=4pt,
    fonttitle=\bfseries,
    sharp corners=south,
    fontupper=\small\ttfamily
  }
}

\begin{document}

% paper title

\title{VERDI: VLM-Embedded Reasoning \\for Autonomous Driving}

% % You will get a Paper-ID when submitting a pdf file to the conference system
% \author{Author Names Omitted for Anonymous Review. Paper-ID 51.}

\author{
    \IEEEauthorblockN{
        Bowen Feng\textsuperscript{*}\IEEEauthorrefmark{2},
        Zhiting Mei\textsuperscript{*}\IEEEauthorrefmark{2},
        Julian Ost\IEEEauthorrefmark{2},
        Filippo Ghilotti\IEEEauthorrefmark{3},
         Baiang Li\IEEEauthorrefmark{2}, \\
        Roger Girgis\IEEEauthorrefmark{3},
        Anirudha Majumdar\IEEEauthorrefmark{2},
        Felix Heide\IEEEauthorrefmark{2}\IEEEauthorrefmark{3}
    }
    \IEEEauthorblockA{
    % \IEEEauthorrefmark{1}Equal Contribution\quad
    \IEEEauthorrefmark{2}Princeton University\quad
    \IEEEauthorrefmark{3}Torc Robotics}
    \IEEEauthorblockA{{\href{https://verdi-driving.github.io}{\textbf{{verdi-driving.github.io}}}}} %TODO: add url
}

% \author{Author Names Omitted for Anonymous Review.}% Paper-ID 18}

\maketitle
\pagestyle{plain}

\footnotetext[1]{* indicates equal contribution.}

\input{Sections/abstract}    

\IEEEpeerreviewmaketitle

\input{Sections_arxiv_new/introduction}

\input{Sections_arxiv_new/relatedWork}

\input{Sections_arxiv_new/method}
\input{Sections_arxiv_new/experiments}
\input{Sections_arxiv_new/conclusion}
\quad

\section*{Acknowledgments}
Zhiting Mei and Anirudha Majumdar were partially supported by the NSF CAREER Award \#2044149, the Office of Naval Research (N00014-23-1-2148), and a Sloan Fellowship. Felix Heide was supported by an NSF CAREER Award \#2047359, a Packard Foundation Fellowship, a Sloan Research Fellowship, a Sony Young Faculty Award, a Project X Innovation Award, and an Amazon Science Research Award. The authors would like to thank Mario Bijelic for helpful discussions, paper editing and proofreading on this work. 

\quad

%% Use plainnat to work nicely with natbib. 
% \bibliographystyle{plainnat}
% \bibliographystyle{unsrtnat}
% \balance
\bibliographystyle{IEEEtran}
\bibliography{references}
% \input{Arxiv_main.bbl}

% Appendix
\appendices
\input{Sections_arxiv_new/appendix}
\end{document}

%% file: cvpr/preamble.tex
\newcommand{\todo}[1]{{\color{red}#1}} %% Bowen: For the editing
\newcommand{\TODO}[1]{\textbf{\color{red}[TODO: #1]}}
\newcommand*{\algname}{\textsc{VERDI}}
\renewcommand{\TODO}[1]{}
\renewcommand{\todo}[1]{#1}

\usepackage{float}

\newcommand{\ECCV}[1]{\textcolor{black}{#1}}

\usepackage{booktabs} 
\usepackage{graphicx}
\usepackage{makecell}

%% file: Sections/abstract.tex
\begin{abstract}
     %Autonomous driving (AD) stacks rely heavily on precise perception of the surrounding environment for downstream planning and navigation. However, partial observability of the real world, such as occlusion and limited sensor range, poses fundamental limits for perception models considered in isolation. Human drivers are capable of leveraging commonsense reasoning to make near-optimal decisions with limited information. Recent work has attempted to leverage the human-like reasoning capabilities of Vision-Language Models (VLMs) at inference time. However, although existing methods are successful in benchmark evaluations, the most successful VLMs are impractical to deploy. Specifically, VLMs of size \todo{give concrete number and size of parameters} require \todo{XX runtime and memory}, which is restrictive for real-time decision making, and their monolithic network structure prohibits safety decomposition. 
     While autonomous driving (AD) stacks struggle with decision making under partial observability and real-world complexity, human drivers are capable of applying commonsense reasoning to make near-optimal decisions with limited information. Recent work has attempted to leverage finetuned Vision-Language Models (VLMs) for trajectory planning at inference time to emulate human behavior. Despite their success in benchmark evaluations, these methods are often impractical to deploy (a 70B parameter VLM inference at merely 8 tokens per second requires more than 160G of memory), and their monolithic network structure prohibits safety decomposition.
     To bridge this gap, we propose \textbf{V}LM-\textbf{E}mbedded \textbf{R}easoning for autonomous \textbf{D}r\textbf{I}ving (\textbf{\algname{}}), a training-time framework that distills the reasoning process and commonsense knowledge of VLMs into the AD stack. \algname{} augments modular differentiable end-to-end (e2e) AD models by aligning intermediate module outputs at the perception, prediction, and planning stages with text features explaining the driving reasoning process produced by VLMs. By encouraging alignment in latent space, \algname{} enables the modular AD stack to internalize structured reasoning, without incurring the inference-time costs of large VLMs. 
     \ECCV{We evaluate \algname{} in both open-loop and closed-loop settings. Our method outperforms existing end-to-end approaches without embedded reasoning by up to \todo{11\%} in $\ell_{2}$ distance, and achieves the best overall driving performance in the closed-loop HugSim simulator, including \todo{a 10\% improvement in Non-Collision Rate}, while maintaining fast inference speed.}
\end{abstract}

%% file: Sections_arxiv_new/introduction.tex
\section{Introduction}
\label{sec:introduction}
Real-world autonomous vehicle deployments have drastically increased in the last three years~\cite{di2024comparative,jung2024safety, chen2022learning, liang2022effective, zeng2019end, zhang2022beverse, zhang2023learning}. However, a major bottleneck to achieving safe and reliable driving arises from the complexity of real-world environments. Dynamic agents, including pedestrians, cyclists, and human-driven vehicles, along with their interactions, present significant challenges for autonomous vehicles to perceive, predict, and plan. Decision-making under semantic ambiguity, partial observability, and unforeseen so-called edge case scenarios remain a critical challenge for autonomous systems~\cite{zhang2024oatracker}. 

 Highly performant autonomous driving (AD) stacks rely on modularized differentiable end-to-end (e2e) architectures~\cite{hu2023planning, jiang2023vad, chen2024vadv2, li2024ego}. Although existing methods are successful in benchmark evaluations, they are often limited to the training datasets consisting of human driving trajectories that lack access to the underlying reasoning processes of the driver~\cite{caesar2020nuscenes, sun2020scalability}. As such, conventional supervised e2e models lack semantic understanding of the driving process, which hinders their real-world applications in unseen scenarios~\cite{hwang2024emma}.
 On the other hand, human drivers respond to challenging decisions and navigate through ambiguities by interpreting visual context and reasoning. In addition, they leverage commonsense knowledge and prior experience to make satisfactory decisions even with incomplete information~\cite{simon1990bounded, jones1999bounded}. This motivates the research question:
%Similarly, autonomous driving stacks should possess comparable reasoning abilities, harnessing human-like commonsense to achieve a deeper semantic understanding of visual data, enabling rapid and robust decision-making in unforeseen and complex situations. This prompts an important research question: 
\emph{How can modular AD models leverage human-like decision-making processes with reasoning and commonsense knowledge when confronted with the real-time challenges of driving?} 

A growing body of work investigates Vision-Language Models (VLMs) in driving tasks for reasoning~\cite{hwang_emma_2024, xing_openemma_2024, chen2024driving, xu2024drivegpt4, wang2024omnidrive, sima2024drivelm, wang2023drivemlm}. Specifically, researchers have augmented AD datasets with VLM annotations to incorporate scene descriptions and explanations~\cite{tian2024drivevlm, wang2023drivemlm, sima2024drivelm}, questions and answers~\cite{wang2024omnidrive, chen2024driving}, and driving rationales~\cite{hwang2024emma, xing_openemma_2024, sima2024drivelm}. The augmented datasets are then used to train or fine-tune VLMs to be multi-modal foundation models that are capable of generating future behaviors for autonomous vehicles, along with natural language explanations for interpretability.
However, querying fine-tuned multi-modal VLMs at inference time could be prohibitively slow. For example, on an OrinX chip, a 4B Qwen model can decode a 1,078-token prompt at 44.5 tokens per second~\cite{tian2024drivevlm}. This latency renders them impractical for real-world safety-critical scenarios~\cite{hwang2024emma}.
To circumvent this, prior work resorts to hierarchical planning and control, using small VLMs, or optimizing the model~\cite{tian2024drivevlm}. 
Moreover, many existing VLM methods forgo multi-modal perception and prediction modules and directly output actions from input queries, making safety decomposition a challenge and allowing for more VLM hallucinations at inference time; we illustrate an instance of this in Appendix~\ref{app:inference_time_vlm_hallucination}. 

%An increasing number of methods leverage Vision-Language Models (VLMs) in autonomous driving tasks\todo{more citations here}. Existing approaches typically use VLMs eiter as generalist policies directly~\cite{hwang_emma_2024, wang_drivecot_2024} or as specialized sub-task solvers, such as for perception.\todo{three more citations here} However, all these approaches query the VLM at inference time, introducing significant latency often up to several seconds—that makes them impractical for real-world, safety-critical scenarios. Those VLM based methods that rely heavily on chain-of-thought prompting and extensive multimodal reasoning for precise answering, although powerful, have yet to address how to efficiently transfer learned capabilities to practical real-world driving without incurring prohibitive inference overhead. How to embedding VLM reasoning into simpler, lightweight architectures to avoid latency issues remains largely unexplored.

\begin{figure*}[t]
    % \vspace{-35pt}
    \centering  %%%was width = 0.81
    \includegraphics[width=0.9\linewidth]{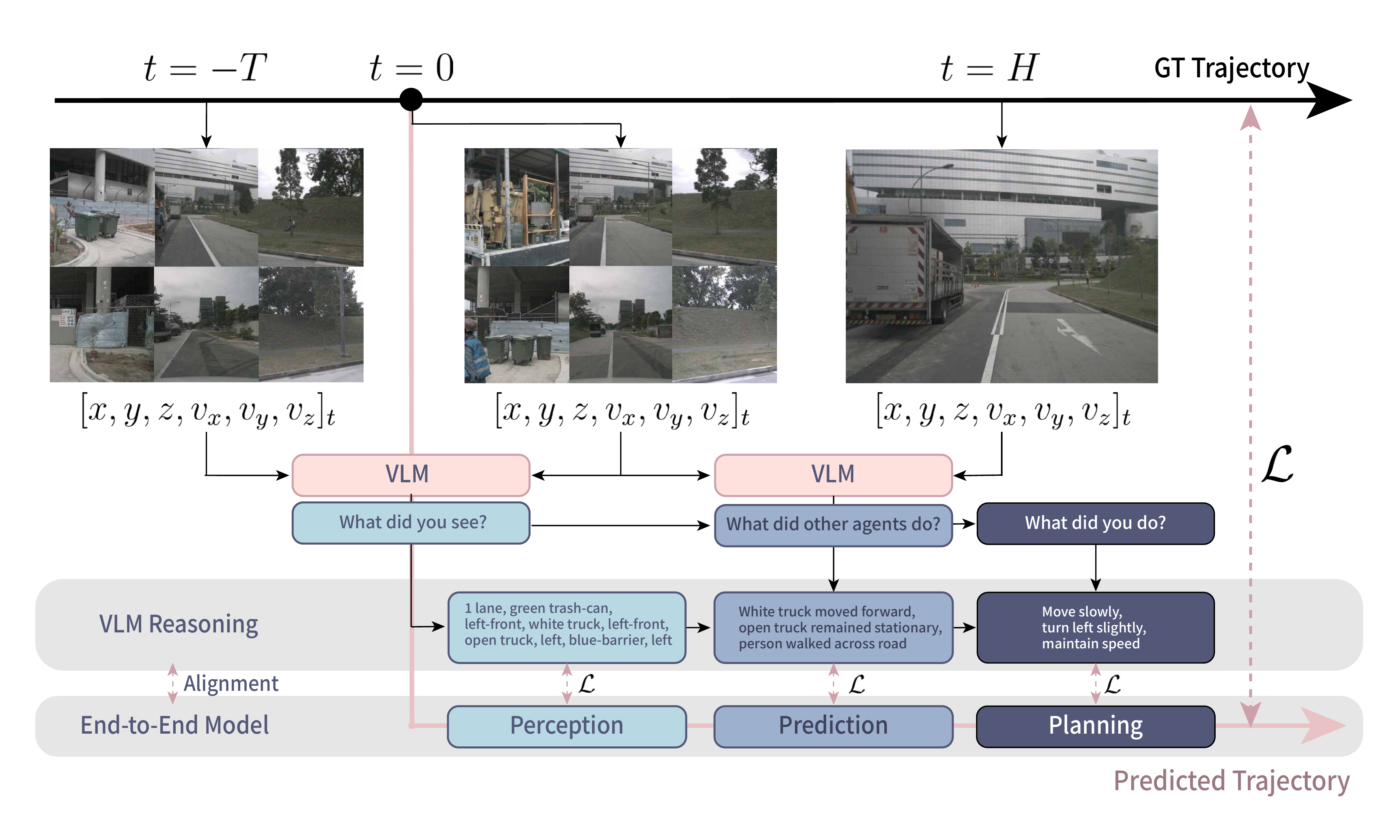}
    % \vspace{-13pt}
    \caption{\textbf{Overview of \textsc{VERDI}.} Our pipeline aligns the VLM reasoning module with our e2e driving model. During training, the ground truth (GT) trajectory and observed images are provided to the VLM for it to explain the reasoning throughout perception, prediction, and planning during the driving process. The VLM's answers to each submodule is aligned with the corresponding submodule outputs from the e2e driving model, effectively distilling VLM knowledge and reasoning into the e2e model. During inference time, the e2e model plans future trajectory with embedded reasoning process, without having to query the VLM (pink arrow).}
    \vspace{-15pt}
    \label{fig:pipeline}
\end{figure*}

We introduce VLM-Embedded Reasoning for autonomous DrIving (\textsc{VERDI}) as an approach for 
%which builds upon the VAD framework \cite{jiang2023vad} and 
distilling latent reasoning capabilities from VLMs; Figure~\ref{fig:pipeline} provides an overview of \textsc{VERDI}. 
%During training, LRMA takes as inputs six surrounding-view images, the ego vehicle's future footprints, and Chain-of-Thought (CoT) textual prompts. 
During training, the method operates on two parallel models for a given trajectory: (1) a differentiable e2e driving model that processes current ego state and corresponding sensor inputs (multi-view images) to generate future trajectories, and (2) a latent reasoning pipeline that extracts the reasoning process of the ground-truth future trajectory by querying a VLM. The driving pipeline sequentially handles perception, prediction, and planning subtasks through the differentiable modules. In this work, we choose to build upon VAD~\cite{jiang2023vad} to create our e2e driving pipeline. The reasoning pipeline prompts the VLM using chain-of-thought~\cite{wei2022chain} for each subtask component.
%Those subtasks are connected sequentially through the passage of queries generated by upstream modules. Specifically, the perception module outputs queries consumed by the prediction module, whose resulting queries subsequently guide the planning module to produce the final ego trajectory plan. 
To distill the reasoning capabilities and commonsense knowledge from VLMs to the driving model, \textsc{VERDI} performs latent feature alignment between the driving and reasoning pipelines. %Queries from upstream driving subtasks are projected into a shared latent space. Similarly, VLM-generated CoT texts are encoded into the same latent space. Within this latent space, the latent 
This alignment process aims to match the driving model latent embeddings with the embeddings of the reasoning process, facilitating the transfer of semantic reasoning from large VLMs to real-time driving model. 
At inference time, the reasoning pipeline is no longer required, since it has been effectively distilled into the driving pipeline. This enables \textsc{VERDI} to perform real-time inference while preserving human-like reasoning abilities.
%After completing this distillation process during training, LRMA does not require the VLM during inference. Instead, inference is conducted solely with the lightweight e2e driving model, offering significantly faster inference speed while retaining comparable reasoning capabilities. 

Our main contributions are three-fold:
% \begin{itemize}[noitemsep, topsep=0pt]
%     \item We introduce Chain-of-Thought (CoT) prompting to elicit semantic understanding, prediction, and planning insights from Vision-Language Models (VLMs) based on visual input.
    % \item 
(1) we develop a modular, differentiable, end-to-end autonomous driving (AD) model featuring latent planning, 
(2) we introduce a method to align individual driving subtasks with corresponding natural language in the latent space, and
(3) \ECCV{we evaluate our framework on both open-loop and closed-loop benchmarks, including nuScenes~\cite{caesar2020nuscenes} and HugSim~\cite{zhou2024hugsim} driving simulator. In open-loop settings, \algname{} outperforms baseline methods in $\ell_{2}$ planning distance by \todo{11\%} on the nuScenes. In closed-loop, \algname{} exhibits the highest driving score among all baseline methods and further improves \todo{10\%} over the direct baseline on non-collision rate. Further, we confirm through a series of ablation experiments that our method has successfully distilled VLM reasoning processes in each submodule, and elicits qualitative improvements in driving behavior.}
% \end{itemize}

% In this work, we propose to combine the benefits of both modularized e2e architectures and VLMs, by training the e2e model to reason through VLM supervision. 

% Figure~\ref{fig:pipeline} summarizes our framework. Our method works with any modularized e2e model that trains perception, prediction, and planning modules. For each module, we align the e2e model output features with text features generated by VLM through a chain-of-thought reasoning process~\cite{wei2022chain}. This training procedure effectively distills the VLM's reasoning process and commonsense knowledge into the multi-modal e2e model. Our method outperforms baseline on the NuScenes dataset~\cite{caesar2020nuscenes} by \may{TODO: results}. 

%% file: Sections_arxiv_new/relatedWork.tex
% \vspace{-3pt}
\section{Related Work}%\vspace{-3pt}
%Our work builds on the foundation of end to end autonomous driving, vision-language model in motion planning, knowledge distillation and latent representation alignment, which we review in the following. 

% \input{Sections/fig_cot}

\textbf{Modularized End to End Autonomous Driving}. 
%End-to-end autonomous driving has been extensively investigated over the last few years, with the primary focus on directly learning the mapping from raw sensory inputs to control action signals~\cite{chen2024end, chib2023recent}.
Although autonomous driving stacks were built modular for a long time, end-to-end optimization is a novel problem \cite{chen2024end, chib2023recent}. These emerging approaches minimize information loss incurred in independently optimized modules, where errors accumulate and compound throughout the perception, prediction, and planning processes.
However, attempts to condense the driving task into single monolithic perception to planning end-to-end models lack the level of interpretability and generalizability desired by AD applications, leading to the rise of jointly trainable end-to-end frameworks with intermediate outputs~\cite{karkus2023diffstack, hu2023_uniad, jiang2023vad, chen2024vadv2}. 
%accumulated throughout the process compared to traditional modularized driving frameworks, where 
Driving innovations arise from advances in perception~\cite{shao2023safety, yang2024visual}, prediction~\cite{yang2024behavior, girgis2021latent}, and planning~\cite{renz2022plant, biswas2024quad} modules, while differentiable losses spanning the entire pipeline, exemplified by UniAD~\cite{hu2023planning} and SY-P3~\cite{hu2022st}, serve to connect and unify these components. %, unify these modules, using differentiable losses across the entire processing pipeline.
%Building on this fully differentiable training approach, VAD~\cite{jiang2023vad} and VADv2~\cite{chen2024vadv2} introduce a lightweight perception module through a vectorized encoding strategy, achieving a more efficient scene representation.
Improvements in the representation space covering vectorized encoding strategies in VAD~\cite{jiang2023vad, chen2024vadv2}, splitting dynamic and static environments \cite{DualAD}, parallelization~\cite{weng2024drive} and inter modular dependencies as the ego-status~\cite{li2024ego} have further enhanced performance.
%BEV-Planner~\cite{li2024ego} and PARA-Drive~\cite{weng2024drive} have explored different learning pipeline designs, ego status information, and head losses via this modularized end-to-end approach. 
However, these models primarily rely on supervised learning from human driving demonstration datasets~\cite{caesar2020nuscenes}, lacking human reasoning capabilities significantly restricts their effectiveness in challenging driving scenarios.
%, particularly emergent behaviors and out-of-distribution testing cases.

\textbf{Vision Language Models (VLMs) in Autonomous Driving} are of great interest as they provide 
%With their broad commonsense knowledge, 
advanced reasoning capabilities, and rich semantic scene understandings. %VLMs~\cite{achiam2023gpt, team2023gemini, yang2024qwen2} have led to growing interest in being used to enhance autonomous driving with interpretability and reasoning~\cite{yang2023llm4drive}. 
Specifically, methods such as DriveGPT4~\cite{xu2024drivegpt4}, OmniDrive~\cite{wang2024omnidrive}, Driving with LLM~\cite{chen2024driving}, DriveMLM~\cite{sima2024drivelm, wang2023drivemlm} enrich existing datasets with fine-grained annotations and fine-tune or train VLMs to reason about the scene through textual responses and predict higher-order actions. EMMA~\cite{hwang2024emma} pushes this further by a full integration of perception, road graphs, and control trajectories in an end-to-end fashion, achieving state-of-the art performance.
%In autonomous driving, for instance, DriveGPT4 \cite{xu2024drivegpt4} utilizes LLMs to explain vehicle actions and predict future control signals through interactive communication with the driver. Similarly, OmniDrive \cite{wang_drivecot_2024} employs a 3D vision-language model architecture aimed to improve reasoning and planning performance. Furthermore, DriveVLM \cite{tian2024drivevlm} and EMMA \cite{hwang_emma_2024} leverage chain-of-thought reasoning frameworks to effectively manage modular driving tasks. 
%These methods notably also aid the interpretability of decision-making processes and planned control signals. %Notably, EMMA has achieved state-of-the-art performance on the NuScenes benchmark.
%However, the using large VLMs suffer from long inference times, and face significant challenges in practical deployment, necessitating workarounds (e.g., using the VLM module for low-frequency driving commands and an e2e model for high-frequency controls, use smaller VLMs, in particular less than 4B parameters~\cite{tian2024drivevlm}) that hinder performance. On the other hand, recent approach VLP~\cite{pan2024vlp} uses VLMs during training time only focusing on contrastive alignment between perception representations and language embeddings. In contrast, VERDI exploits the full language reasoning process across perception, prediction, and planning, and aligns it with the corresponding e2e module. 
However, large VLMs cause slow inference and deployment hurdles, forcing compromises, e.g. restricting them to low‐frequency planning while an end-to-end model handles high-frequency controls or capping the size below 4B parameters~\cite{tian2024drivevlm}, which degrades overall performance. %VLP~\cite{pan2024vlp} avoids this by leveraging VLMs only during training via contrastive visual–language alignment, whereas 
% We note that VLP and VLM-AD are closed-source methods, with approaches distinctively different from VERDI. 
\ECCV{Two recent closed-source models, VLP~\cite{pan2024vlp} and VLM-AD~\cite{xu2024vlm}, explore use of VLM during training time with distinctively different approaches. VLP uses language models only for embedding rigid textual descriptions to vectors, disregarding reasoning capabilities in the LM itself. VLM-AD only supervises planning outcome with VLM text annotations, while \algname distills the \textit{reasoning} process and commonsense to e2e submodules, validated in our ablation experiments in Sec.~\ref{subsec:ablations}.
Our approach harnesses language reasoning across perception, prediction, and planning, directly integrating it into the e2e pipeline, and does not require a VLM at runtime. Alpamayo~\cite{wang2025alpamayo} is another recent reasoning-based end-to-end driving VLA trained on human driving demonstrations augmented with reasoning annotations, providing a strong baseline for comparison. In contrast to Alpamayo’s VLA architecture, \algname{} adopts a differentiable modular model in which driving reasoning capabilities are distilled from a VLM during training. This design enables model-based driving with more than 10$\times$ faster inference while maintaining strong performance. We validate this comparison on the HugSim closed-loop benchmark, demonstrating both improved driving efficiency and competitive overall driving performance relative to Alpamayo.}
% While we take inspiration from VLM-AD, our approach is fundamentally different in that we distill the process of the reasoning rather than the result of reasoning (ground-truth actions).

% However, VLM-AD only uses a VLM to annotate the planned behavior, whereas we show that while planning is already supervised by the ground-truth trajectories, aligning the commonsense reasoning process in the modules before planning is more effective.
%However, all these approaches face significant challenges in practical deployment, due to the latency introduced by inferencing with VLMs or LLMs online in the decision-making loop.

\textbf{Knowledge Distillation, Representation Learning and Alignment}.
%Knowledge distillation is a technique to compress and accelerate models by learning a student network, and has gained wide popularity due to the growing size of deep learning models~\cite{hinton2015distilling, gou2021knowledge}. Especially in the era of large language models, knowledge distillation becomes a crucial methodology for transferring knowledge from large models to smaller counterparts, as well as improving the model performance~\cite{xu2024survey}.
%Our work can be interpreted as a form of knowledge distillation, although we do not train a student network to emulate the VLM's behavior. Instead, we train the modularized e2e network to emulate the reasoning processes of VLMs,
Knowledge distillation compresses and accelerates models by training a smaller “student” to mimic a larger “teacher,” and has become essential as deep learning models continue to grow in size~\cite{hinton2015distilling, gou2021knowledge}. In the era of large language models, it is particularly critical for transferring knowledge to more compact architectures and for boosting overall performance~\cite{xu2024survey}. Our approach can be seen as a specialized form of distillation: rather than having a student network emulate the VLM’s outputs directly, we train our modular end-to-end network to replicate the VLM’s internal reasoning processes.
%In the realm of knowledge distillation for AVs, closest to our work is VLM-AD~\cite{xu2024vlm}, where the authors use the last layer features from an end-to-end autonomous driving model, and append additional attention heads to leverage distilled VLM knowledge. However, VLM-AD only uses VLM to annotate the planned behavior, whereas we show that while planning is already supervised by the ground-truth trajectories, distilling the commonsense reasoning process in the modules leading up to planning is more effective.
%\textbf{Representation Learning and Alignment}.
%by aligning representations from task-specific modules with VLM text features. 
This multi-modal aligned representation learning draws insight from contrastive learning~\cite{radford_learning_2021, cherti_reproducible_2023} and embedding space supervision~\cite{khosla2020supervised}.
% CLIP~\cite{radford_learning_2021, cherti_reproducible_2023} and similar models such as Sentence-Transformers~\cite{reimers2019sentence} that have demonstrated how large-scale pretraining on aligned image-text or text-text pairs can produce powerful and transferable representations.
%Contrastive learning objectives, in particular, have proven effective in shaping semantic structure in latent space and enabling cross-modal grounding. 
Recently,~\cite{sancaktar2025sensei} has explored leveraging these pretrained embeddings by aligning them with learned modules for downstream control tasks, such as active exploration.
%For example, Sensei~\cite{sancaktar2025sensei} proposes to use VLM-derived rewards to encourage active exploration by distilling an ``interestingness'' prior. KD-VLP~\cite{liu2021kd} improves vision-language pretraining by using object-level alignment between vision tokens and text phrases to guide masked modeling. 
%Our method builds on distillation and representation alignment, distilling VLM reasoning process to the driving model by aligning text and driving representations.
Our method combines knowledge distillation with cross‐modal alignment, embedding VLM reasoning into the driving model by aligning its textual and driving feature representations.

%While these methods primarily use representation-level supervision, our work aligns intermediate activations and submodules in a goal-directed system. This enables more interpretable and composable policy learning, rooted in the abstraction and reasoning capabilities of foundation models.\todo{Why? Claim modularity here?}

%% file: Sections_arxiv_new/method.tex
\section{VLM-Embedded Reasoning}
\label{sec:method}
% \vspace{-3pt} 
\subsection{Problem Formulation} 
We cast the autonomous driving problem as future trajectory planning. Let $s_t=[x, y, z, v_x, v_y, v_z]_t$ be the ego agent state at time step $t$. The history trajectory of the past $T$ time steps is denoted as $\tau_{-T:0}:=\{s_{-T},...,s_0\}$, and the future trajectory with a planning horizon $H$ is denoted $\tau_{0:H}:=\{s_0, ..., s_H\}$. We have image observations at each time step $t$, denoted $o_t$. We aim to learn an end-to-end model $\mathcal P_\theta$, that plans future trajectories given only sequential observations:
\begin{equation}
    \hat\tau_{0:H}\sim\mathcal P_\theta(\tau_{0:H}|o_{-T:0}),
\end{equation}
where $\theta$ represents learnable weights of the model. As shown in Figure~\ref{fig:pipeline}, our training pipeline focuses on the alignment of two modules: the VLM reasoning module, which we denote $\mathcal M$, and the e2e driving model, which we denote $\mathcal P$. Further, we assume that the modular e2e model can be broken down into three submodules $\mathcal P_\theta^i$, where $i\in\{\texttt{perception}, \texttt{prediction}, \texttt{planning}\}$. In addition to the standard losses for AD (e.g., $\ell_2$ distance), we add a training objective to minimize the difference between all VLM reasoning steps and the intermediate e2e submodules. Our key idea is to accomplish this by aligning the text features from the VLM reasoning module, $f_{\mathcal M}$, with the driving features from the e2e model, $f_{\mathcal P}$, through a similarity loss in the overall loss function, which we describe in Section~\ref{subsec:e2e}.
We detail how we obtain language embeddings in Section~\ref{subsec:prompt}, how we augment the e2e model by alignment with VLM embeddings in Section~\ref{subsec:e2e}.

%%%%%%%%%%%%%%%%Move elsewhere%%%%%%%%%%%%%

% For each module, we utilize the human-like reasoning outputs of the VLM and align the e2e model in addition to the ground truth information as follows.

% Just like for the VLM, we project output features $\mathcal P_{i,\theta_i} \left( o \right)$ at the end of each module $i$ into the latent space, shared with the VLM output projections. For this transformation we introduce our Progressive Feature Projector (PFP) $\phi_i$ with the learnable parameters $\phi_i$ at the end of each module, that results in the projected e2e feature $f_{\mathcal P_i} = \text{PFP}_{\phi_i}\left( F_i \right)$  and  $F_i = \mathcal P_{i,\theta_i} \left( * \right)$, where $*$ represents the input to each module $i$.

% We then use cosine similarity to represent feature alignment loss between the e2e features and the language features:
% \begin{equation}
%     \mathcal L_{f}(f_{\mathcal P}, f_{\mathcal M}) = \frac{f_{\mathcal P}\cdot f_{\mathcal M}}{\Vert f_{\mathcal P}\Vert \Vert f_{\mathcal M}\Vert }.
% \end{equation}
% The total loss for each module $i$ is then:
% \begin{equation}
%     \mathcal L_i = \mathcal L_e (\theta_i) + \lambda\mathcal L_{f}(f_{\mathcal P}, f_{\mathcal M}),
% \end{equation}
% where $\lambda$ is a weight hyperparameter balancing the two sources of losses and ${\mathcal L}_e(\theta_i)$ represents the loss function on ground-truth annotations and data according to the respective e2e model.

%%%%%%%%%%%%Move elsewhere%%%%%%%%%%%%%

\subsection{Obtaining Language Embeddings}
\label{subsec:prompt}

\input{Sections/fig_cot}

In this section, we describe how we acquire language features $f_{\mathcal M}$ from the VLM reasoning module $\mathcal M$. This module consists of two steps: (1) querying the VLM to obtain text responses, and (2) mapping the text response to latent features.
Figure~\ref{fig:cot} outlines the VLM reasoning module.

\textbf{Prompting Strategy.} Following the modular e2e architecture, we prompt the VLM to reason about the perception, prediction, and planning steps in a chain-of-thought fashion~\cite{wei2022chain}. We first guide the VLM with some system prompt to ensure its responses are concise and relevant. 
For perception, the VLM is asked to explain the past trajectory. It is given multi-view images captured from six cameras at timesteps $t=-T$ and $t=0$, as well as the past ego trajectory $\tau_{-T:0}$. The VLM is then asked to identify the number of lanes in the scene, any agents or notable objects present, as well as their relative locations. To prevent spatial confusion and redundancy, the VLM is asked to label the agents and objects from left to right and from front to back.
For prediction and planning, the VLM is asked to explain the future trajectory. In both queries, it is given the future front images at $t=0$ and $t=H$, as well as the future trajectory $\tau_{0:H}$. For prediction, the answer from perception is provided and the VLM is asked to describe other agents' actions in the same order they were listed. The planning query in addition takes the prediction answers, and asks the VLM to describe what actions the ego agent took. The full prompt is listed in Appendix~\ref{app:implementation_details}.

\textbf{Encoding VLM Responses.} We then map the answer for each module to latent feature space, using a text encoder. The text encoder is able to map natural language to semantically meaningful sentence embeddings~\cite{reimers2019sentence}. This encoding process reduces the dimensionality of the text, makes it machine interpretable and allows for alignment while preserving semantic meaning. The encoded text results in a feature vector $f_{\mathcal M}^i$ for each module $i$.

\subsection{Model Alignment with VLM Reasoning}
\label{subsec:e2e}

\input{Sections/fig_training}

To obtain driving features $f_{\mathcal P}^{i}$ for each module $i$ of the e2e model, we project its output features $F_i = \mathcal P^{i}_{\theta} \left( \cdot \right)$ into the shared latent space with $f_{\mathcal M}^i$ via a learnable Progressive Feature Projector (PFP) $\phi_i$, producing the e2e feature $f_{\mathcal P}^{i} = \text{PFP}{\phi_i}\left( F_i \right)$ (more details in Appendix~\ref{app:implementation_details}). Figure~\ref{fig:training} shows the training architecture that aligns the e2e model and VLM reasoning.

% \begin{figure*}[t!]
%     \vspace{-20pt} 
%     \centering
%     \includegraphics[width=0.9\linewidth]{figures/E2E-Training-Framework.pdf}
%     \vspace{-8pt}
%     \caption{\textbf{\textsc{VERDI} Training.} The e2e model is trained with \textsc{VERDI} for the individual perception, prediction, and planning modules. All relevant feature maps $F$ and $Q$ are first mapped to a feature $f_{\mathcal P}$ in a representation space, which is shared with the encoded language features $f_{\mathcal{M}}$. This mapping is facilitated by \textsc{VERDI}'s trainable $\text{PFP}$ layers. The perception outputs $F_{\texttt{perception}}$ including the extracted image features, are directly supervised with the encoded VLM features. In the subsequent modules, all features are supervised. $\mathcal{L}_f$ computes their similarity. %between the language and e2e features.
%     }
%     \vspace{-15pt} 
%     \label{fig:training}
% \end{figure*}

% We project output features $F^{i}$ from each module $i$, into the latent space, shared with the VLM output projections. For this projection, we introduce our Progressive Feature Projector (PFP) $\phi_i$ with the learnable parameters $\phi_i$ of each module, that results in the projected e2e feature $f^{i} = \text{PFP}^{i}_{\phi}\left( F_i \right)$  and  $F_i = \mathcal P_{\theta_i} \left( * \right)$, where $*$ represents the input to each module $i$.
% \subsubsection{VLM Reasoning Alignment}
% \label{subsec:alignment}
\textbf{VLM-Distilled Scene Understanding.} To enhance the precision of the Bird's Eye View (BEV) map's representation of the driving scene, we find it essential to extract high-quality visual features $F_{\texttt{perception}}$. Vision-Language Models (VLMs), trained on large-scale internet image-text pairs, possess strong capabilities in identifying object classes, road markings, and spatial relationships with respect to the ego vehicle. Our goal is to distill this semantic understanding from the VLM into the image features $F_{\texttt{perception}}$ encoded by the image feature encoder, as presented in Figure~\ref{fig:training}. 
The features are then processed by the BEV Encoder to produce a refined BEV representation. 
To achieve alignment, we project $F_{\texttt{perception}}$ to $f_{\mathcal{ P}}^{\texttt{perception}}$ in the shared latent space with $\text{PFP}^{\texttt{perception}}_{\phi}$, which consists of a combination of CNN and MLP layers(See Appendix~\ref{app:implementation_details} for details). The perception driving features are then ready to be aligned with VLM reasoning features.
%\todo{To achieve this, we leverage the VLM's textual response generated from raw multi-view images as $I_{-T}$ and $I_{0}$, and align it with the intermediate image features $F_{\texttt{perception}}$ before they are processed into the BEV encoder as shown in \ref{fig:training} Image Feature Extractor. $\text{PFP}^{\texttt{perception}}_{\phi}$, a combination of CNN and MLP layers, projects $F_{\texttt{perception}}$ from e2e space into the shared latent space.}

% However, a fundamental challenge lies in the disparity between the feature spaces of the VLM's text output and the visual features. To address this, we designed the $\textbf{PFP}_{perception}$ with a combination of CNN and MLP. The detailed implementation can be found in the Appendix. 

% \textbf{VLM-Distilled Agents Prediction.} In general, the e2e prediction module performs the prediction from the BEV features $F_{bev}$ for a horizon over the next $H$ timesteps. In figure\ref{fig:training}, $F_{bev}$ is decoded into agent and map queries, $Q_{agents}$ and $Q_{map}$, which are then concatenated to form the unified prediction query:
% \begin{equation}
%     Q_{predict} = [Q_{agents}; Q_{map}],
% \end{equation}
\textbf{VLM-Distilled Agents Prediction.} Next, the end‐to‐end prediction module forecasts the next $H$ timesteps from the BEV feature map $F_{\texttt{bev}}$. In Figure~\ref{fig:training}, $F_{\texttt{bev}}$ is decoded into agent and map queries, $Q_{\texttt{agents}}$ and $Q_{\texttt{map}}$, which presents a common architecture among e2e models~\cite{jiang2023vad, chen2024vadv2, hu2023_uniad}. These two separate feature maps are then concatenated to form the unified prediction query
\[
Q_{\texttt{predict}}
\;=\;
\bigl[\,Q_{\texttt{agents}};\,Q_{\texttt{map}}\,\bigr],
\]
where $[\,\cdot\,;\,\cdot\,]$ denotes channel-wise concatenation. Here, $Q_{\texttt{agents}}$ encodes agent-agent relative positions and $Q_{\texttt{map}}$ encodes agent-map spatial relations.
% \todo{where $Q_{\texttt{agents}}$ and $Q_{\texttt{map}}$ describes the latent embedding of the agent-agent relative position and agent-map relative position, and ``;'' denotes the concatenation operation. }

To imbue the prediction module with the VLM’s reasoning ability, we distill the VLM's reasoning ability by aligning VLM's prediction $f^{\texttt{prediction}}_{\mathcal{M}}$ and $f^{\texttt{prediction}}_{\mathcal{P}}$, which we obtain by projecting $Q_{\texttt{predict}}$ through an MLP-based $\text{PFP}_{\phi}^{\texttt{prediction}}$.

\textbf{VLM Distilled Ego Planning.} We feed the combined query $Q_{\texttt{predict}}$, which integrates both other agents' future trajectories and spatial map context, into the planning module $\mathcal P_{\texttt{planning}}$. Conditioned on $Q_{\texttt{predict}}$, the planner produces the latent ego feature $Q_{\texttt{ego}}$ as:
\begin{equation}
    Q_{\texttt{ego}} = \mathcal P^{\texttt{planning}}_\theta(\tau_{0:H}|Q_{\texttt{predict}}).
\end{equation}

We project $Q_{\texttt{ego}}$ onto the latent space with an MLP based $\text{PFP}_{\phi}^{\texttt{planning}}$, before passing it into the last-layer decoder. 
This allows us to enrich $Q_{\texttt{ego}}$ with VLM reasoning by aligning VLM's $f^{\texttt{planning}}_{\mathcal M}$ with the latent e2e planning features $f^{\texttt{planning}}_{\mathcal P}$.

\textbf{Alignment Loss}
% \label{subsec:loss}
We use cosine similarity to represent the feature alignment loss $\mathcal L_{f}(f_{\mathcal P}, f_{\mathcal M})$ between the e2e features and the language features, \par
% \vspace{-5pt}
% \begin{center}
% \begin{minipage}{.5\linewidth}
\begin{equation}
    \mathcal L_{f}(f_{\mathcal P}, f_{\mathcal M}) = \frac{f_{\mathcal P}\cdot f_{\mathcal M}}{\Vert f_{\mathcal P}\Vert \Vert f_{\mathcal M}\Vert }\text{, with a total loss}
\end{equation}
\par
% \end{minipage}
% \begin{minipage}{.45\linewidth}
\begin{equation}
    \mathcal L_i = \mathcal L_e (\theta_i) + \lambda_i\mathcal L_{f}(f_{\mathcal P}^i, f_{\mathcal M}^i),
\end{equation}
% \end{minipage}%
% \end{center}
% \vspace{-5pt}
% \begin{equation}
%     \mathcal L_{f}(f_{\mathcal P}, f_{\mathcal M}) = \frac{f_{\mathcal P}\cdot f_{\mathcal M}}{\Vert f_{\mathcal P}\Vert \Vert f_{\mathcal M}\Vert }.
% \end{equation}
% The total loss for each module $i$ is then
% \begin{equation}
%     \mathcal L_i = \mathcal L_e (\theta_i) + \lambda_i\mathcal L_{f}(f_{\mathcal P}^i, f_{\mathcal M}^i),
% \end{equation}
where ${\mathcal L}_e(\theta_i)$ represents the supervised  loss function with the original ground-truth data according to \cite{jiang2023vad} and $\lambda$ is a weight hyperparameter balancing the two contributions. \ECCV{In the following experiments, we select the $\lambda = 10$ so that all loss terms are balanced in the same order of magnitude.}

\begin{table*}[!h]
    \vspace{-16pt}
    \centering
    \caption{\textbf{Evaluation on the nuScenes dataset} \cite{caesar2020nuscenes}. Methods are compared according to: (1) Whether a VLM is required at inference; (2) Inference speed (FPS); (3) Trajectory accuracy, measured as the $\ell_{2}$ distance to the expert trajectory at 1s, 2s, and 3s horizon; and (4) Whether precise historical ego‐vehicle state is used in planning. In a fair comparison with methods not having privileged access to ego status, including location, \algname{} achieves the best performance across all metrics. \ECCV{The best results are \textbf{bolded}}.}
     \resizebox{0.9\linewidth}{!}{ %% was 0.8
    \begin{tabular}{lccccccc}
    \toprule
         Method& \makecell{\small Requires VLM \\ \small  @ Inference} & FPS $\uparrow$ & $l_2$ (1s) $\downarrow$ &  $l_2$ (2s) $\downarrow$ &  $l_2$ (3s) $\downarrow$ &  $l_2$ (avg.) $\downarrow$  &  Ego Status \\
    \midrule
    \midrule
         DriveVLM~\cite{tian2024drivevlm}& $\checkmark$& 2.43& 0.18&  0.34&  0.68&  0.40 &   $\checkmark$\\
         OpenEMMA~\cite{xing_openemma_2024}& $\checkmark$& NA& 1.45& 3.21& 3.76& 2.81 & $\checkmark$\\
        \midrule
         OmniDrive~\cite{wang2024omnidrive}& $\checkmark$& 0.44&1.15& 1.96& 2.84& 1.98 & -\\
         %VLP~\cite{pan2024vlp}& NA& 0.26& 0.47& 0.78& 0.50 && $\checkmark$\\
         %VLM-AD~\cite{xu2024vlm}& NA& 0.24& 0.46& 0.75& 0.48 &GPT-4o& -\\
         UniAD~\cite{hu2023_uniad}& - & 1.8& 0.48&  0.96&  1.05&  0.83 &    -\\
         VAD-Base~\cite{jiang2023vad}& - &\textbf{4.5}& 0.41&  0.70&  1.05&  0.72 &   -\\
         \textsc{VERDI}& - &\textbf{4.5}&\todo{\textbf{0.36}} & \todo{\textbf{0.62}}& \todo{\textbf{0.96}}& \todo{\textbf{0.65}} & -\\
    \bottomrule
    \end{tabular} }
    \label{tab:main_results}
    % \vspace{-15pt}
\end{table*}

\begin{table*}[!t]
\centering
\caption{\ECCV{\textbf{Evaluation on the HugSim Closed-loop Simulator Benchmark} \cite{zhou2024hugsim} with the HUGSIM nuScenes dataset. 
    Metrics are averaged by difficulty levels. The best results between architectural-equivalent models (VAD, \textsc{VERDI}) are \textbf{bolded}. We further show the best results overall as \underline{underlined}. We report scene-level performance with core metrics, including No Collision (NC), Drivable Area Compliance (DAC), Time-To-Collision (TTC), Comfort (COM), Route Completion (Rc), and HugSim Driving Score (HDScore). 
    \textsc{VERDI} achieves the \textbf{best overall HDScore} and improves NC by $10\%$ over VAD-Base, \emph{without requiring VLM inference}. 
    \textbf{Efficiency.} Each entry reports total parameters (Milions) and average latency per planning prediction (milliseconds) on a single GPU.}\label{tab:hugsim_benchmark}}

\resizebox{\textwidth}{!}{%
\begin{tabular}{@{\hspace{2mm}}c@{\hspace{4mm}} l c c | l c c c c c c c}
\toprule
% group col intentionally blank
& \textbf{Method}
& \makecell{\small \textbf{Requires VLM} \\ \small \textbf{@ Inference}}
& \makecell{\small \textbf{Params/Latency} \\ \small \textbf{[M]/[ms]}}
& \textbf{Difficulty}
& \textbf{NC} $\uparrow$ & \textbf{DAC} $\uparrow$ & \textbf{TTC} $\uparrow$
& \textbf{COM} $\uparrow$ & \textbf{Rc} $\uparrow$ & \textbf{Overall HDScore} $\uparrow$ \\
\midrule\midrule

% =========================================================
% Different architecture: 4 methods × 5 rows = 20
% =========================================================
\multirow{20}{*}{\rotatebox{90}{\textbf{Different Architectures}}}
& \multirow{5}{*}{\ECCV{Alpamayo~\cite{wang2025alpamayo}}}
& \multirow{5}{*}{$\checkmark$}
& \multirow{5}{*}{11079 / 5246}
& easy    & 0.976 & 1.000 & 0.969 & 0.922 & 0.046 & 0.044 \\
&  &  &  & medium  & 0.977 & 1.000 & 0.969 & 0.921 & 0.046 & 0.043 \\
&  &  &  & hard    & 0.968 & 1.000 & 0.963 & 0.906 & 0.041 & 0.038 \\
&  &  &  & extreme & 0.962 & 1.000 & 0.955 & 0.891 & 0.042 & 0.039 \\
&  &  &  & \textit{Overall} & \underline{0.971} & \underline{1.000} & \underline{0.964} & 0.910 & 0.044 & 0.041 \\
\cline{2-11}\noalign{\vskip 0.6mm}

& \multirow{5}{*}{OpenEMMA~\cite{xing_openemma_2024}}
& \multirow{5}{*}{$\checkmark$}
& \multirow{5}{*}{7000 / 573}
& easy & 0.552 & 0.764 & 0.480 & 1.000 & 0.450 & 0.209 \\
&  &  &  & medium & 0.512 & 0.790 & 0.448 & 1.000 & 0.376 & 0.172 \\
&  &  &  & hard & 0.312 & 0.823 & 0.261 & 1.000 & 0.310 & 0.080 \\
&  &  &  & extreme & 0.388 & 0.805 & 0.350 & 1.000 & 0.291 & 0.104 \\
&  &  &  & \textit{Overall} & 0.441 & 0.796 & 0.385 & \underline{1.000} & 0.357 & 0.141 \\
\cline{2-11}\noalign{\vskip 0.6mm}

& \multirow{5}{*}{OmniDrive~\cite{wang2024omnidrive}}
& \multirow{5}{*}{$\checkmark$}
& \multirow{5}{*}{6849 / 554}
& easy & 0.830 & 0.197 & 0.899 & 0.000 & 0.069 & 0.010 \\
&  &  &  & medium & 0.786 & 0.157 & 0.767 & 0.000 & 0.087 & 0.004 \\
&  &  &  & hard & 0.757 & 0.200 & 0.731 & 0.000 & 0.075 & 0.005 \\
&  &  &  & extreme & 0.766 & 0.224 & 0.686 & 0.000 & 0.074 & 0.006 \\
&  &  &  & \textit{Overall} & 0.785 & 0.194 & 0.771 & 0.000 & 0.076 & 0.006 \\
\cline{2-11}\noalign{\vskip 0.6mm}

& \multirow{5}{*}{UniAD~\cite{hu2023_uniad}}
& \multirow{5}{*}{-}
& \multirow{5}{*}{132 / 724}
& easy & 0.813 & 0.967 & 0.661 & 0.111 & 0.757 & 0.396 \\
&  &  &  & medium & 0.708 & 0.959 & 0.412 & 0.199 & 0.389 & 0.140 \\
&  &  &  & hard & 0.716 & 0.978 & 0.398 & 0.109 & 0.370 & 0.113 \\
&  &  &  & extreme & 0.649 & 0.950 & 0.375 & 0.096 & 0.291 & 0.107 \\
&  &  &  & \textit{Overall} & 0.721 & 0.963 & 0.462 & 0.129 & \underline{0.452} & 0.189 \\

% \cline{1-11}\noalign{\vskip 0.8mm}
% \cline{1-11}\noalign{\vskip 0.8mm}
\midrule
\midrule
% =========================================================
% Architecturally equivalent: 2 methods × 5 rows = 10
% =========================================================
\multirow{10}{*}{\rotatebox{90}{\textbf{Equivalent Architecture}}}
& \multirow{5}{*}{VAD-Base~\cite{jiang2023vad}}
& \multirow{5}{*}{-}
& \multirow{5}{*}{58 / 359}
& easy & 0.768 & 0.880 & 0.611 & 0.897 & 0.499 & 0.314 \\
&  &  &  & medium & 0.537 & 0.912 & 0.305 & 0.897 & 0.308 & 0.108 \\
&  &  &  & hard & 0.631 & 0.877 & 0.387 & 0.934 & 0.367 & 0.170 \\
&  &  &  & extreme & 0.450 & 0.957 & 0.320 & 0.876 & 0.328 & 0.154 \\
&  &  &  & \textit{Overall} & 0.597 & 0.906 & 0.406 & 0.901 & \textbf{0.375} & 0.186 \\
\cline{2-11}\noalign{\vskip 0.6mm}

& \multirow{5}{*}{VERDI}
& \multirow{5}{*}{-}
& \multirow{5}{*}{58 / 359}
& easy & 0.797 & 0.945 & 0.664 & 0.963 & 0.470 & 0.328 \\
&  &  &  & medium & 0.649 & 0.927 & 0.466 & 0.958 & 0.315 & 0.143 \\
&  &  &  & hard & 0.623 & 0.941 & 0.389 & 0.951 & 0.368 & 0.213 \\
&  &  &  & extreme & 0.566 & 0.961 & 0.358 & 0.944 & 0.265 & 0.115 \\
&  &  &  & \textit{Overall} & \textbf{0.659} & \textbf{0.944} & \textbf{0.469} & \textbf{0.954} & 0.354 & \underline{\textbf{0.200}} \\
\bottomrule
\end{tabular}}
\vspace{-15pt}
\end{table*}

%% file: Sections/fig_cot.tex
\begin{figure*}[t!]
    % \vspace{-28pt} 
    \centering
    \includegraphics[width=0.95\linewidth]{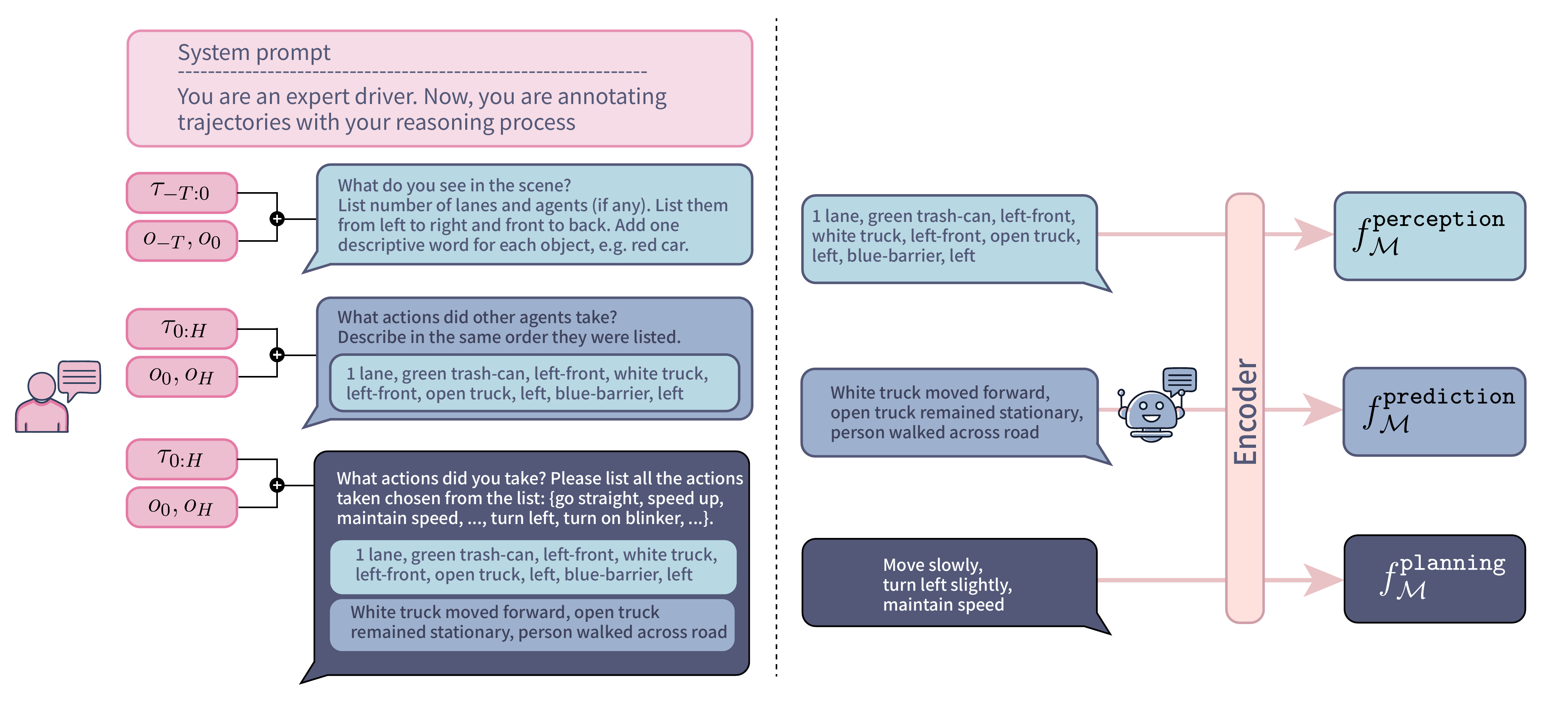}
    % \vspace{-12pt} 
    \caption{Obtaining description features through chain-of-thought prompting and text encoder. For each query, the prompt consists of the system prompt, the observed images, the ego vehicle trajectory, the respective question, as well as the answers to the upstream modules (if any). The VLM answers to each module are encoded and mapped to a latent feature space.}
    \label{fig:cot}
    \vspace{-25pt} 
\end{figure*}

%% file: Sections/fig_training.tex
\begin{figure*}[t!]
    % \vspace{-20pt} 
    \centering
    \includegraphics[width=\linewidth]{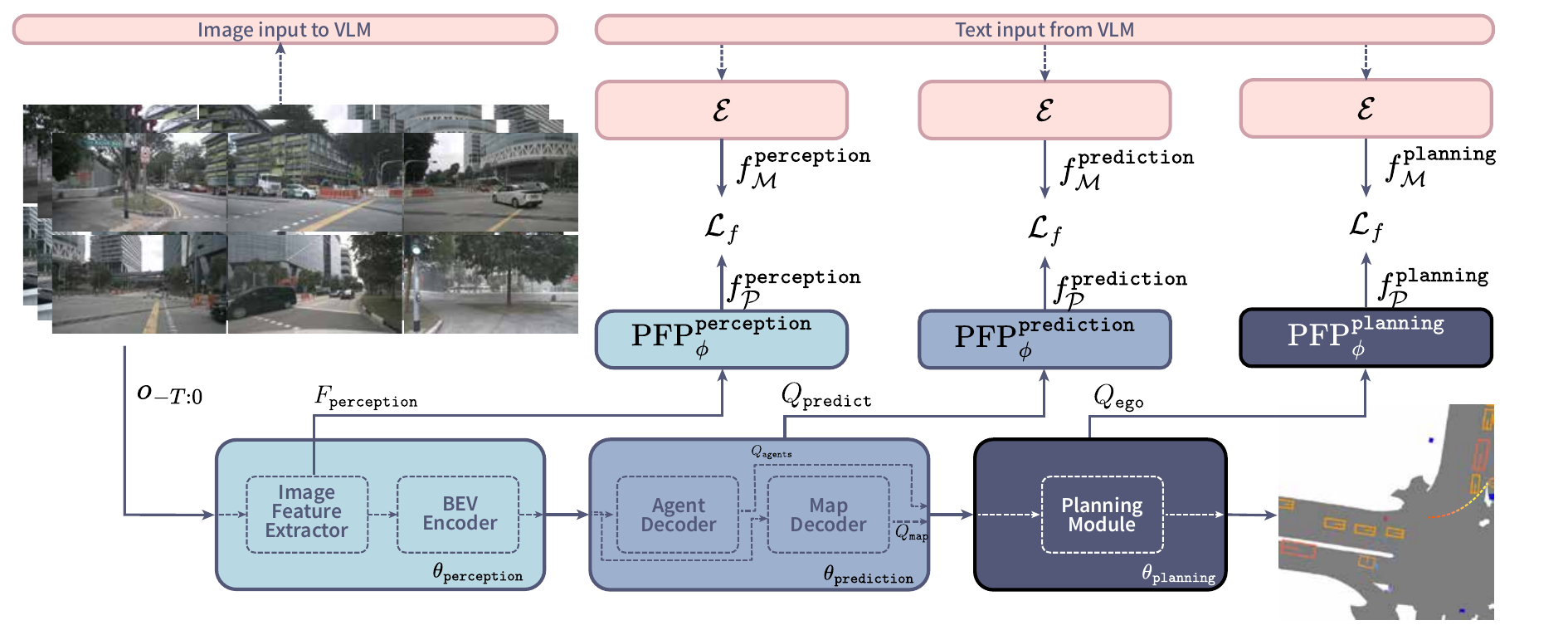}
    % \vspace{-8pt}
    \caption{\textbf{\textsc{VERDI} Training.} The e2e model is trained with \textsc{VERDI} for the individual perception, prediction, and planning modules. All relevant feature maps $F$ and $Q$ are first mapped to a feature $f_{\mathcal P}$ in a representation space, which is shared with the encoded language features $f_{\mathcal{M}}$. This mapping is facilitated by \textsc{VERDI}'s trainable $\text{PFP}$ layers. The perception outputs $F_{\texttt{perception}}$ including the extracted image features, are directly supervised with the encoded VLM features. In the subsequent modules, all features are supervised. $\mathcal{L}_f$ computes their similarity. %between the language and e2e features.
    }
    \vspace{-15pt} 
    \label{fig:training}
\end{figure*}

%% file: Sections_arxiv_new/experiments.tex
\section{Experiments}
\label{sec:experiments}
We confirm the efficacy of \algname{} through a series of experiments on several driving benchmarks, both open-loop and closed-loop, comparing against baseline approaches in Section~\ref{subsec:main-results}, and validating all design choices with ablation experiments in Section~\ref{subsec:ablations}. 

\textbf{Baseline Approaches.}
We compare our method to two main categories of baselines: (1) e2e methods that directly train a modular differentiable network for future trajectory planning, (2) methods that train or finetune a VLM to infer future actions or control signals. 
We implement our method building on the e2e model VAD-Base~\cite{jiang2023vad}, making it a direct baseline. We also compare against UniAD~\cite{hu2023planning} , a representative state-of-the-art e2e method. For VLM-based methods, we compare against DriveVLM~\cite{tian2024drivevlm}, which generates chain-of-thought reasoning and hierarchical plans using a VLM, OmniDrive~\cite{wang2024omnidrive}, which aligns 3D BEV features with counterfactual scene reasoning via a fine-tuned VLM, OpenEMMA~\cite{xing_openemma_2024}, an open-source implementation of Waymo's EMMA, predicting full planner trajectories using a fine-tuned VLM and AlphaMayo\cite{wang2025alpamayo}, a reasoning-augmented driving VLA that integrates large language model–based semantic reasoning with end-to-end perception and planning in long-tail driving scenarios.

\textbf{Dataset and Metrics.}
% \cvpr{We evaluate \algname{} across a number of challenging driving benchmarks under both open- and closed-loop settings, including nuScenes~\cite{caesar2020nuscenes},  Bench2Drive dataset~\cite{jia2024bench2drive} collected in the Carla Simulator~\cite{dosovitskiy2017carla}, and HugSim Simulator~\cite{zhou2024hugsim}.
\ECCV{We evaluate \algname{} across challenging driving benchmarks under both open- and closed-loop settings, including nuScenes~\cite{caesar2020nuscenes} and HugSim Simulator~\cite{zhou2024hugsim}.}

% \textbf{Open-Loop Evaluations.} We conduct experiments on the nuScenes dataset~\cite{caesar2020nuscenes}, which contains approximately 1000 driving scenes with 150 validation and 850 trainin sequences, each about 20 seconds long. The dataset includes 6 camera views, and provides annotated keyframes at 2Hz. For planning evaluation, we follow the standard open-loop protocol~\cite{hu2023_uniad}: given 2 seconds of past observations, the model predicts the vehicle's trajectory over the next 3 seconds. Performance is measured using $\ell_2$ displacement errors at 1, 2, and 3-second horizons. We also compare inference speed in frames per second (FPS) for applicable methods. Following the same evaluation protocol, we conduct our experiments to the Bench2Drive dataset~\cite{jia2024bench2drive}, which includes roughly 10,000 driving scenes derived from the CARLA~\cite{dosovitskiy2017carla} driving simulator. The training scenes are
% recorded from six camera views and annotated at 10 Hz.
\ECCV{\textbf{Open-Loop Evaluations.} We conduct experiments on the nuScenes dataset~\cite{caesar2020nuscenes}, which contains approximately 1000 driving scenes with 150 validation and 850 training sequences, each about 20 seconds long. The dataset includes 6 camera views, and provides annotated keyframes at 2Hz. For planning evaluation, we follow the standard open-loop protocol~\cite{hu2023_uniad}: given 2 seconds of past observations, the model predicts the vehicle's trajectory over the next 3 seconds. We evaluate the model using $\ell_2$ displacement errors at 1, 2, and 3-second horizons. We also compare inference speed in frames per second (FPS) for applicable methods.} 
% Following the same evaluation protocol, we conduct our experiments to the Bench2Drive dataset~\cite{jia2024bench2drive}, which includes roughly 10,000 driving scenes derived from the CARLA~\cite{dosovitskiy2017carla} driving simulator. The training scenes are
% recorded from six camera views and annotated at 10 Hz.

\textbf{Closed-Loop Evaluations.} To further assess the robustness and effectiveness of \algname{} in realistic and interactive environments, we perform closed-loop evaluations on the
%both the CARLA~\cite{dosovitskiy2017carla} and 
HugSim~\cite{zhou2024hugsim} driving simulator. The HugSim closed-loop benchmark includes 88 test scenes derived from the nuScenes testing dataset, featuring much denser traffic and challenging vehicle interactions.
%and a greater number of surrounding vehicles than the original nuScenes scenes, resulting in a more interactive and challenging driving scenario. Trained on the Bench2Drive dataset, VERDI is further evaluated in the CARLA simulator across 220 scenes. Both simulator benchmarks assess
Driving performance is assessed through a diverse set of sub-metrics beyond collision and goal-reaching, including Drivable Area Compliance (DAC) and Comfort (COM) in HugSim, 
%as well as Emergency Braking and Overtaking in Bench2Drive, 
providing a more comprehensive and safety-oriented evaluation of driving behavior. 

% \textbf{Implementation Details.} 
% \cvpr{For the nuScenes dataset, training was conducted for 60 epochs on 10 NVIDIA A6000 GPUs with a batch size of 2 using the AdamW optimizer.
% All other training configurations maintain the settings adopted from VAD~\cite{jiang2023vad}.
% % TODO: Supplementary with a cosine learning‐rate scheduler.
% Both open loop benchmark inference were done on a single NVIDIA A6000 GPU for all models. Closed-loop HugSim inference was done on a single NVIDIA A10G.
% For the Bench2Drive dataset, training was operated on the full dataset, with finetuning for 6 epochs on the released VAD checkpoint.} For VLM language embeddings, we use \texttt{Qwen-2.5-VL-72B}~\cite{yang2024qwen2} to generate reasoning for every trajectory in the nuScenes training dataset. \cvpr{Considering the scale of the Bench2Drive training set and the substantial time required, we employ \texttt{Qwen-2.5-VL-7B}~\cite{yang2024qwen2} to generate reasonings on the Bench2Drive training clips.} We use chain-of-thought prompting strategy described in Section~\ref{subsec:prompt} to generate text outputs from multi-view images and states, and project the text back into an embedding space using the \texttt{all-mpnet-base-v2} sentence-transformers model~\cite{noauthor_sentence-transformersall-mpnet-base-v2_2024, reimers2019sentence}.
\textbf{Implementation Details.} 
\ECCV{For the nuScenes dataset, training was conducted for 60 epochs on 10 NVIDIA A6000 GPUs with a batch size of 2 using the AdamW optimizer.
All other training configurations maintain the settings adopted from VAD~\cite{jiang2023vad}.
% TODO: Supplementary with a cosine learning‐rate scheduler.
Both open loop benchmark inference were done on a single NVIDIA A6000 GPU for all models. Closed-loop HugSim inference was done on a single NVIDIA A10G.
For VLM language embeddings, we use \texttt{Qwen-2.5-VL-72B}~\cite{yang2024qwen2} to generate reasoning for every trajectory in the nuScenes training dataset.} 
% For the Bench2Drive dataset, training was operated on the full dataset, with finetuning for 6 epochs on the released VAD checkpoint.} For VLM language embeddings, we use \texttt{Qwen-2.5-VL-72B}~\cite{yang2024qwen2} to generate reasoning for every trajectory in the nuScenes training dataset. 

% \cvpr{Considering the scale of the Bench2Drive training set and the substantial time required, we employ \texttt{Qwen-2.5-VL-7B}~\cite{yang2024qwen2} to generate reasonings on the Bench2Drive training clips.} We use chain-of-thought prompting strategy described in Section~\ref{subsec:prompt} to generate text outputs from multi-view images and states, and project the text back into an embedding space using the \texttt{all-mpnet-base-v2} sentence-transformers model~\cite{noauthor_sentence-transformersall-mpnet-base-v2_2024, reimers2019sentence}.

% \vspace{-5pt}
\subsection{Quantitative Assessment}
\label{subsec:main-results}
\subsubsection{Open-Loop Evaluations}
Table~\ref{tab:main_results} summarizes our quantitative experimental findings. Note that we separately evaluate methods that condition trajectory planning on \emph{history ego status -- privileged information that typically perform  better~\cite{li2024ego} -- and real-time deployable methods that do not have access to the precise ego status.} Among baselines that do not use ego status, VERDI achieves the lowest average $\ell_2$ distance to the expert trajectory \todo{(0.65)}—an \todo{$10\%$} improvement over our direct baseline, VAD-Base. While DriveVLM—leveraging ego status—attains a lower $\ell_2$ distance, VERDI runs $\sim 1.9$ times faster. Compared to baseline methods that report inference speed (e.g., $2.44$ FPS for DriveVLM~\cite{tian2024drivevlm}, $0.44$ FPS for OmniDrive implemented with Pytorch~\cite{wang2024omnidrive}). VERDI offers the best trade-off between inference speed and trajectory accuracy among methods without ego status.

\subsubsection{Closed-Loop Evaluations}
Table~\ref{tab:hugsim_benchmark} reports closed-loop experimental results. \ECCV{\algname{} achieves the best overall driving performance with the least model parameters and the inference time latency.}Compared to the direct baseline VAD-Base, \algname{} achieves approximately $8\%$ improvement in overall driving performance and a $10\%$ higher No-Collision (NC) rate, demonstrating that enhanced reasoning capability leads to safer driving behavior in the same architecture. \ECCV{We observe a small drop in Route Completion (Rc) compared to VAD-Base and confirm that this is due to the safe early stopping behavior of \algname, where VAD-Base completes slightly longer drives at the expense of collisions. We present additional qualitative analysis in~\Cref{sec:results_qualitative}. 
Notably, VERDI outperforms recent large reasoning-based driving models, Alpamayo (approximately $200\times$ larger and $10\times$ slower), in HDScore: 0.2 (\algname) vs 0.041 (Alpamayo), an approximately $79.5\%$ overall driving performance improvement.}

% \cvpr{Table~\ref{tab:b2d_open} reports the learboard of the Bench2Drive open loop benchmarks~\cite{jia2024bench2drive}. \algname{} performs the best among feature distillation based methods and achieves $11\%$ performance improvement over its direct baseline VAD, demonstrating the effectiveness of the reasoning capability alignment from the VLM with the e2e model.}

\subsection{Qualitative Assessment}
\label{sec:results_qualitative}
\begin{figure*}[t!]
    % \vspace{-22pt}
    \centering
    \includegraphics[width=\linewidth]{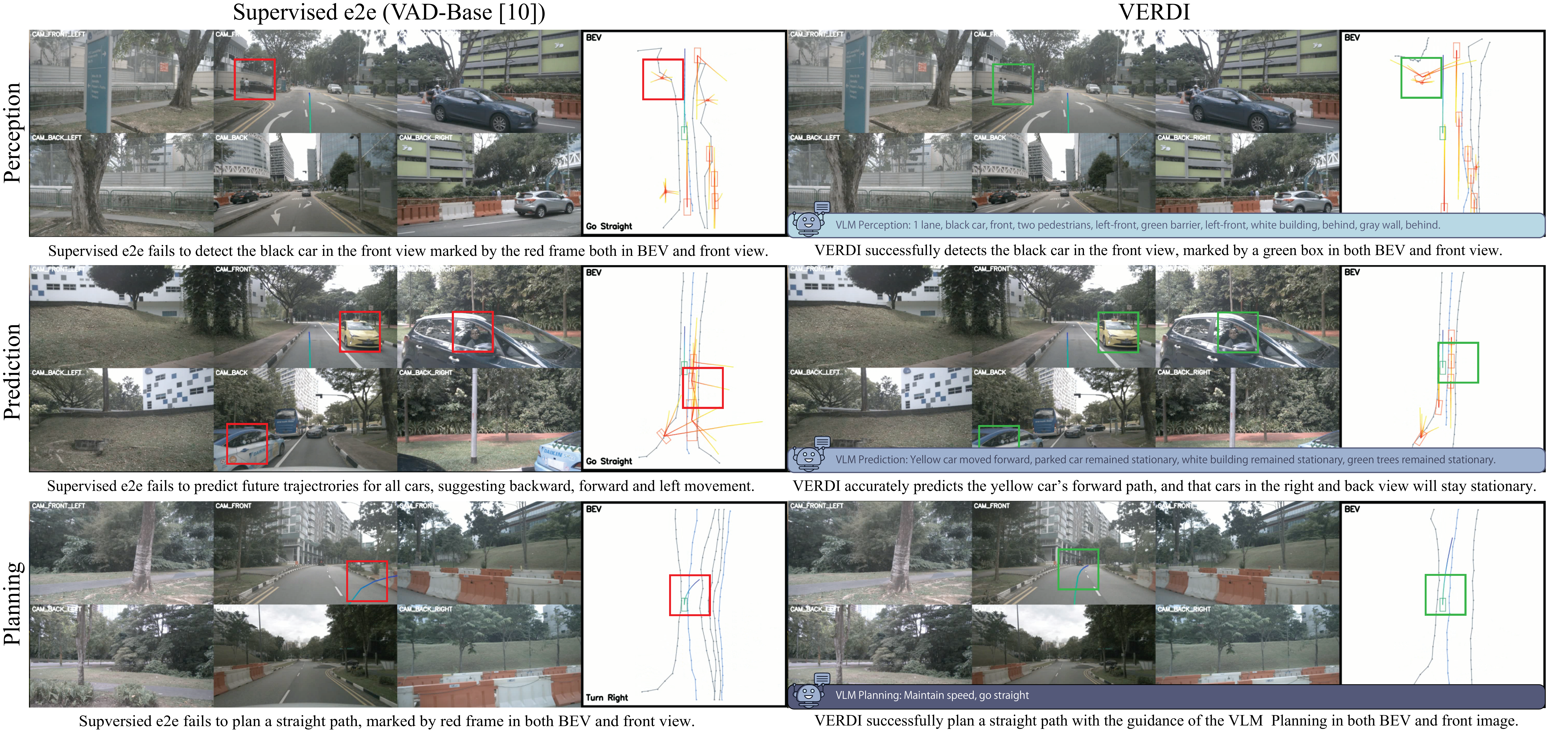}
    % \vspace{-12pt}
    \caption{Qualitative comparison of \algname{} (Ours, right column) and the Supervised e2e model (baseline, left column) on the nuScenes dataset \cite{caesar2020nuscenes}. Each entry shows the multi-view camera observations on the left and the BEV view on the right at one time step $t$. The left panel overlays the ego agent’s planned 3-second trajectory on the front-camera image and BEV panel as a solid green line that fades to blue. The BEV panel renders the ego vehicle as a green rectangle, pedestrians and other vehicles as red rectangles, and their predicted 3-second trajectories as red lines. %The ego agent’s own planned trajectory is projected onto the BEV as a green-to-blue fading line.
    Each example shows our successful performance on the perception, prediction, and planning modules, indicated by \textcolor{green}{$\square$}, while failures are highlighted by \textcolor{red}{$\square$}. We also show the VLM texts response for each case used during training time.}
    \vspace{-15pt}
    \label{fig:qual}
\end{figure*}

We present qualitative comparisons between the Supervised e2e (VAD-Base) (left column) and \algname{} (right column) \ECCV{on nuScenes} in Figure~\ref{fig:qual}. Our approach yields better perception results: for example, in the top row, the supervised e2e detects only the pedestrians with red sparkle sign and fails to localize the front-left vehicle in the BEV view \textcolor{red}{$\square$}, whereas our model correctly identifies both the pedestrians and that vehicle \textcolor{green}{$\square$}. Detecting this vehicle is critical as omitting it could lead to a cross-traffic collision. \algname{} also has more accurate future motion prediction. In the second row, the ego vehicle continues straight, while the front-right taxi, the adjacent black car, and the rear white-and-blue taxi all wait at the traffic light, indicating they will likewise proceed straight once it turns green. \algname{} correctly predicts these straight-ahead trajectories shown as red lines aligned with each vehicle’s heading, whereas the supervised e2e baseline mistakenly forecasts leftward and backward movements. This confirms that the VLM commonsense reasoning substantially improves the accuracy of predictions on other agents’ motion. In the third row, \algname{} produces a mostly straight trajectory with a slight rightward adjustment, as shown in both the front-view and BEV projections. In contrast, the Supervised e2e baseline chooses a sharp right turn that would collide with the barrier.

\begin{figure}[!htbp]
  \centering
    \includegraphics[width=\linewidth]{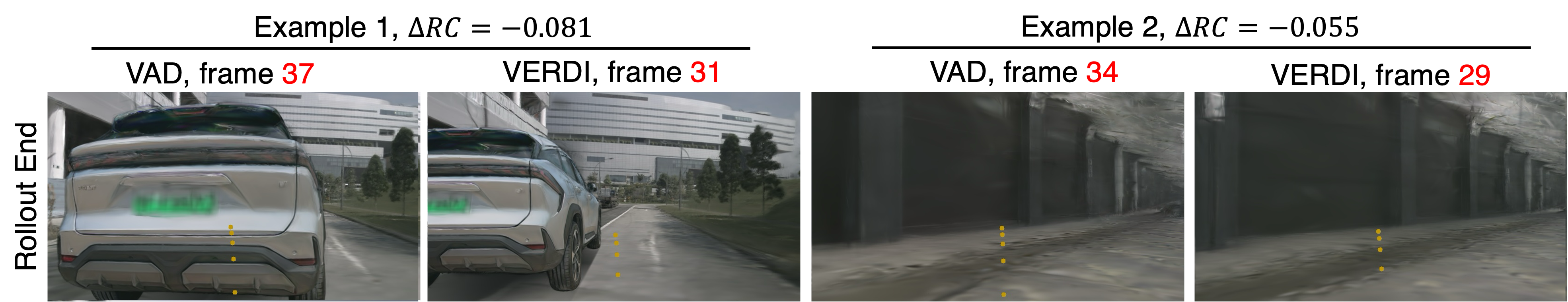}
    % \vspace{-22pt}
    \caption{Example scenes with RC drops. We qualitatively show that decreases in the RC score do not necessarily correspond to meaningful differences in metric distance. In those two examples, the RC drops are $0.081$ and $0.055$, respectively, yet both trajectories result in collisions at the same location.}\label{fig:Qual_CL_Failure}
    % \vspace{-15pt}
\end{figure}

\ECCV{
% In~\Cref{fig:Qual_CL_Failure}, w
We further illustrate the more cautious behavior \algname exhibits across all scenario difficulties compared to VAD-Base, which leads to the small drop in route completion (Rc) rate in~\Cref{tab:hugsim_benchmark}. Since Rc is computed as the percentage of the completed driving distance, a higher Rc (driving longer distances) does not necessarily correspond to better driving behavior. %As highlighted by Table ~\ref{tab:hugsim_benchmark}, wherever VERDI features a lower RC, it outperforms VAD on \emph{all} safety metrics, supporting our hypothesis that it has learned safer driving behaviors. Quantitative analysis shows that 
\algname's Rc is lower than VAD-Base in $19$ out of $80$ total test scenes: $15$ out of $19$ such cases have Time-To-Collision flagged, indicating a high-likelihood of an imminent crash, that VERDI avoided before reaching rollout. Further, we show qualitatively that in the remaining $4$ out of those $19$ cases, there is no significant route execution compromise. In these failure cases for both, \algname collides with the same object only a few frames earlier (\Cref{fig:Qual_CL_Failure}), without a significant difference in metric distance. }

\vspace{-10pt}
\subsection{Ablation Study}\vspace{3pt}
\label{subsec:ablations}
We conduct ablation experiments to analyze the contribution of each component. All model derivations are trained based on VAD-tiny with a sub-sampled set of nuScenes, which we describe in more detail in Appendix C. %Appendix~\ref{app:ablation_loss_and_details}. 
% The set consists of trajectories from each scene of the full dataset starting at $ t=[0, 10, 20]$ and lasting for $10$ timesteps. Since most scenes last around $40$ timesteps the trajectories evenly span the full dataset,

\begin{table}[h] %%%% Original table 4.
\centering
    \caption{\textbf{Ablations Module Alignment}. Results on aligning different e2e modules \ECCV{tested in the nuScenes dataset with $\boldsymbol{\ell_2}$ error metric}. Bolded numbers stand for the best and italic numbers stand for second best.}
    \vspace{1mm}
    \resizebox{\linewidth}{!}{
    \begin{tabular}{ccccccc}
    \toprule
         \multicolumn{3}{c}{\textbf{Aligned Modules}} & \multicolumn{4}{c}{$\boldsymbol{\ell_2}$ \textbf{error} $\downarrow$} \\
    \cmidrule(lr){1-3} \cmidrule(lr){4-7}
    \textbf{Perception} & \textbf{Prediction} & \textbf{Planning} & (1s) & (2s) & (3s) & (avg.) \\
    \midrule
    \midrule
         & &&  1.23&  2.07&  2.94& 2.08\\
         $\checkmark$& && \textbf{1.14}& 1.91& 2.70&1.92\\
         & $\checkmark$&&  1.17&  1.95&  2.77& 1.96\\
         & &$\checkmark$&  1.15&  \textbf{1.90}&  2.70& 1.92\\
         $\checkmark$& $\checkmark$&&  1.19&  1.99&  2.82& 2.00\\
         $\checkmark$& &$\checkmark$&  1.20&  1.97&  2.76& 1.98\\
         & $\checkmark$&$\checkmark$&  1.16&  \textbf{1.90}&  \textbf{2.66}& \textbf{1.90}\\
         $\checkmark$& $\checkmark$&$\checkmark$&  \textbf{1.14}&  \textbf{1.90}&  \textit{2.69}& \textit{1.91}\\
    \bottomrule
    \end{tabular}}
    
    \label{tab:ablation-each-module}
\vspace{-5pt}
\end{table} %%%% Original table 4

\textbf{Alignment Module Variations.} In Table~\ref{tab:ablation-each-module}, we show that each alignment module improves the performance at varying rates. To evaluate models trained with aligning only certain modules $p$, we simply set $\lambda_i=10$ for $i=p$ and $\lambda_j=0$ for $j\neq p$.  % TODO: Move details to appendix!! All models are trained for 30 epochs and evaluated on the full test dataset. 
Notably, aligning only features of the perception module improves $\ell_2$ loss in the short term (1 second), while aligning only the prediction and planning modules improves $\ell_2$ loss in the long term (3 seconds). Aligning with all three achieves a desired balance of short and long-term objectives, resulting in a low average $\ell_2$ loss. %This result is consistent with the way we split the trajectories into past and future to align with each module, causing alignment with only perception to be short-sighted. Aligning with the planning module alone achieves good performance, possibly because planning reasoning uses a query containing perception and prediction data. 
We provide more training details in Appendix~\ref{app:ablation_loss_and_details}, showing the loss curve saturating.

\ECCV{%\textbf{Significance of Prediction/Planning Alignment. } The perception modules shows strong performance improvement in nuScens $\boldsymbol{\ell_2}$ test. Yet, the effectiveness of Prediciton/Planning module on the overall performance is still under investigated. 
To further investigate the effect of aligning the prediction and planning modules on the overall performance, we follow the same protocol to ablate these modules in the HugSim Closed-loop simulator (\cref{tab:p1p2p3}).
% Following the same training protocol as that in nuScenes, we further test the prediction/planning alignment significance in HugSim Closed-loop simulator (\cref{tab:p1p2p3}) as a sanity check. 
Aligning all modules yields the best overall driving performance. Compared with aligning only perception and prediction, planning leads to a $52\%$ driving performance increase. Similarly, incorporating prediction in addition to perception and planning modules alone contributes to a $30\%$ performance gain.} \par
\begin{table}[ht!]
% \vspace{-10pt}
\centering
\caption{\ECCV{Ablation experiments testing aligning perception and prediction modules in the HugSim closed-loop simulator. Following the same protocol as~\cref{tab:ablation-each-module}, we train VAD-tiny with a sub-sampled set of nuScenes. Aligning \emph{all modules} achieves the best HugSim Driving score.}}
\setlength{\tabcolsep}{3pt}
\resizebox{\columnwidth}{!}{
\begin{tabular}{@{}lcccccc@{}}
\toprule
\textbf{Modules} & \textbf{NC} & \textbf{DAC} & \textbf{TTC} & \textbf{COM} & \textbf{Rc} & \textbf{HDScore} \\
\midrule
Perception + Prediction    & 0.612 & 0.746 & 0.529 & 0.188 & 0.103 & 0.027 \\
Prediction + Planning    & 0.538 & 0.708 & 0.406 & 0.990 & 0.107 & 0.039 \\
All  & 0.454 & 0.714 & 0.339 & 0.935 & 0.176 & \textbf{0.056} \\
\bottomrule
\end{tabular}}

\label{tab:p1p2p3}
\vspace{-12pt}
\end{table}

% \begin{table}[t]
% \centering
%     \caption{\textbf{Ablations VLM embeddings}. Ablation results with VLM embeddings of varying quality.}
%     \vspace{1mm} 
%     \resizebox{0.5\linewidth}{!}{
%     \begin{tabular}{lcccc}
%     \toprule
%       \textbf{VLM Embeddings}&  $\ell_2$ (1s) $\downarrow$ &  $\ell_2$ (2s) $\downarrow$&  $\ell_2$ (3s) $\downarrow$& $\ell_2$ (avg.) $\downarrow$\\
%     \midrule
%     \midrule
%          Baseline (no VLM) &  1.23&  2.07&  2.94& 2.08\\
%          Adversarial VLM&  1.22&  2.04&  2.88& 2.05\\
%          QwenVL-7B-CLIP&  1.20&  2.01&  2.84& 2.02\\
%          QwenVL-72B-CLIP&  1.19&  1.99&  2.84& 2.01\\
%          QwenVL-72B-ST-mini& 1.21& 2.04& 2.89&2.05\\
%          QwenVL-72B-ST-base& \textbf{1.15}& \textbf{1.93}& \textbf{2.74}&\textbf{1.94}\\
%          \bottomrule
%     \end{tabular}
%     }
%     \label{tab:ablation-vlm-quality}
% % \end{wraptable}
% \end{table}
\begin{table}[t]
\centering
    \caption{\textbf{Ablation results with VLM embeddings of varying quality.}}
    \vspace{1mm} 
    \resizebox{\linewidth}{!}{
    \begin{tabular}{lcccc}
    \toprule
      \textbf{VLM Embeddings}&  $\ell_2$ (1s) $\downarrow$ &  $\ell_2$ (2s) $\downarrow$&  $\ell_2$ (3s) $\downarrow$& $\ell_2$ (avg.) $\downarrow$\\
    \midrule
    \midrule
         Baseline (no VLM) &  1.23&  2.07&  2.94& 2.08\\
         Zero tensor&  2.47 &  4.01 &  5.46 & 3.98\\
         Random tensor& 3.01 & 4.93 & 6.75 & 4.90\\
         QwenVL-7B-CLIP&  1.20&  2.01&  2.84& 2.02\\
         QwenVL-72B-CLIP&  1.19&  1.99&  2.84& 2.01\\
         QwenVL-72B-ST-mini& 1.21& 2.04& 2.89&2.05\\
         QwenVL-72B-ST-base& \textbf{1.15}& \textbf{1.93}& \textbf{2.74}&\textbf{1.94}\\
         \bottomrule
    \end{tabular}
    }
    \label{tab:ablation-vlm-quality}
% \end{wraptable}
\end{table}

\textbf{VLM Embedding Variations.} We investigate the role of VLM embeddings and ablate our model by providing VLM embeddings of varying qualities. % TODO: Move details to appendix!! For this set of experiments, we use the subset of nuScenes consisting of trajectories starting at $ t=[0, 5, 10]$ and lasting for $10$ timesteps for each scene, and each model is trained for 10 epochs. 
In Table~\ref{tab:ablation-vlm-quality}, we report that using a higher-quality VLM (Qwen-72B rather than Qwen-7B) improves performance. The choice of the text encoder and hence the shared latent space impacts performance. Further, we find that using CLIP~\cite{radford_learning_2021} as text encoder yields worse performance than the \texttt{all-mpnet-base-v2} sentence-transformers model, since bag of words used by CLIP leads to less semantically meaningful embeddings. \ECCV{Moreover, we verify that the performance gains arise from meaningful VLM supervision rather than arbitrary signals. Specifically, we replace the VLM features with random or zero tensors during training. As shown in~\cref{tab:ablation-vlm-quality}, training \algname with random or zero features leads to substantial performance degradation on nuScenes. Using zero features increases the $\ell_2$ error by $+91\%$, while random features result in a $+136\%$ increase.}

% \ECCV{\textbf{Quantitative \& Qualitative Analysis of Rc Drops.} We find that the $2.1\%$ drop in Route Completion (RC) in [56] from VAD-Base to {VERDI} in Table ~\ref{tab:hugsim_benchmark} is caused by {VERDI} exhibiting more cautious behavior across all scenario difficulties. VAD-Base exhibits aggressive and generally less safety-considerate driving behavior. Rc is computed as the percentage of the completed driving distance. As such, driving longer distances does not necessarily correspond to better driving behavior. As highlighted by Table ~\ref{tab:hugsim_benchmark}, wherever VERDI features a lower RC, it outperforms VAD on \emph{all} safety metrics, supporting our hypothesis that it has learned safer driving behaviors. Quantitative analysis shows that our RC is lower than baseline in $19$ out of $80$ total test scenes: $15$ out of $19$ such cases have TTC values flagged, indicating a high-likelihood of an imminent crash, that VERDI avoided before reaching rollout. Further, we show qualitatively that in the other 4 out of those 19 cases RC drops do not account for significant route execution compromise. In these failure cases for both, \algname{} collides with the same object only a few frames earlier (Fig. \ref{fig:Qual_CL_Failure}), without a significant difference in actual metric distance. }

\textbf{Quantitative Evaluation of Perception/Prediction Module-Level Performance.}%
\begin{table}[t]
\centering
\small
\setlength{\tabcolsep}{8pt}
\renewcommand{\arraystretch}{1.15}

\caption{\ECCV{\textbf{Quantitative Evaluation of Perception and Prediction on nuScenes.}} 
VERDI improves map reconstruction quality and reduces orientation and attribute errors, 
consistent with the qualitative results in Fig.~4.}
\label{tab:percep_pred_rebuttal}

\begin{tabular}{lcc}
\toprule
\textbf{Metric} & \textbf{VAD} & \textbf{VERDI (Ours)} \\
\midrule
Chamfer mAP $\uparrow$ & 0.476 & \textbf{0.502} \\
mAOE $\downarrow$      & 0.532 & \textbf{0.481} \\
mAAE $\downarrow$      & 0.233 & \textbf{0.208} \\
\bottomrule
\end{tabular}
% \vspace{-13pt}
\end{table}
\ECCV{In \Cref{tab:percep_pred_rebuttal}, we further confirm gains in perception/prediction module-level metrics, consistent with our qualitative analysis in~\Cref{fig:qual}. For perception, \algname improves %perception, e.g., better 
map reconstruction (Chamfer mAP$\uparrow$ 0.502 (ours) vs. 0.476 (VAD)). For
%and (refers to Figure 4. Perception). 
prediction, we see consistent reductions in mAOE$\downarrow$ (0.481 vs. 0.532) and mAAE$\downarrow$ (0.208 vs. 0.233),
% (mAOE 0.481 vs 0.532; mAAE 0.208 vs 0.233), 
indicating a more accurate prediction of agent orientation and motion attributes. This suggests that \algname better captures the directional and semantic structure of motion intention. Orientation and attribute errors are particularly important for interaction-heavy scenarios (e.g., turning, yielding, stopping), where small false heading errors can lead to large planning errors despite low ADE/FDE.}

%% file: Sections_arxiv_new/conclusion.tex
\section{Conclusion}
% \vspace{-13pt}
\ECCV{We introduce VERDI, a training-time method that distills Vision–Language Model reasoning into a lightweight end-to-end autonomous-driving stack by aligning each submodule with respective VLM reasoning embeddings. 
Extensive experiments and ablation studies on nuScenes and HugSim closed-loop simulator show that \textsc{VERDI} achieves a \todo{11\%} performance gain in $\ell_{2}$ distance, the best across the HugSim benchmark with \todo{10\%} safety improvements over the direct baseline without reasoning embeddings. Although our unoptimized implementation runs at 4.5 Hz, further acceleration is possible through TensorRT optimization and model quantization, leaving substantial room for faster real-time deployment.
\nopagebreak
In future, we plan to extend the framework to self-supervised latent reasoning alignment with feedback and explore reasoning-guided inference using ultra-small VLMs for real-time control.}

\newpage

%% file: Sections_arxiv_new/appendix.tex
\section{Hallucinations for Inference-time VLMs (Sec. 1.0 main)}
\label{app:inference_time_vlm_hallucination}
% \vspace{-10pt}
\begin{figure}[h!]
    \centering
    \includegraphics[width=0.8\linewidth]{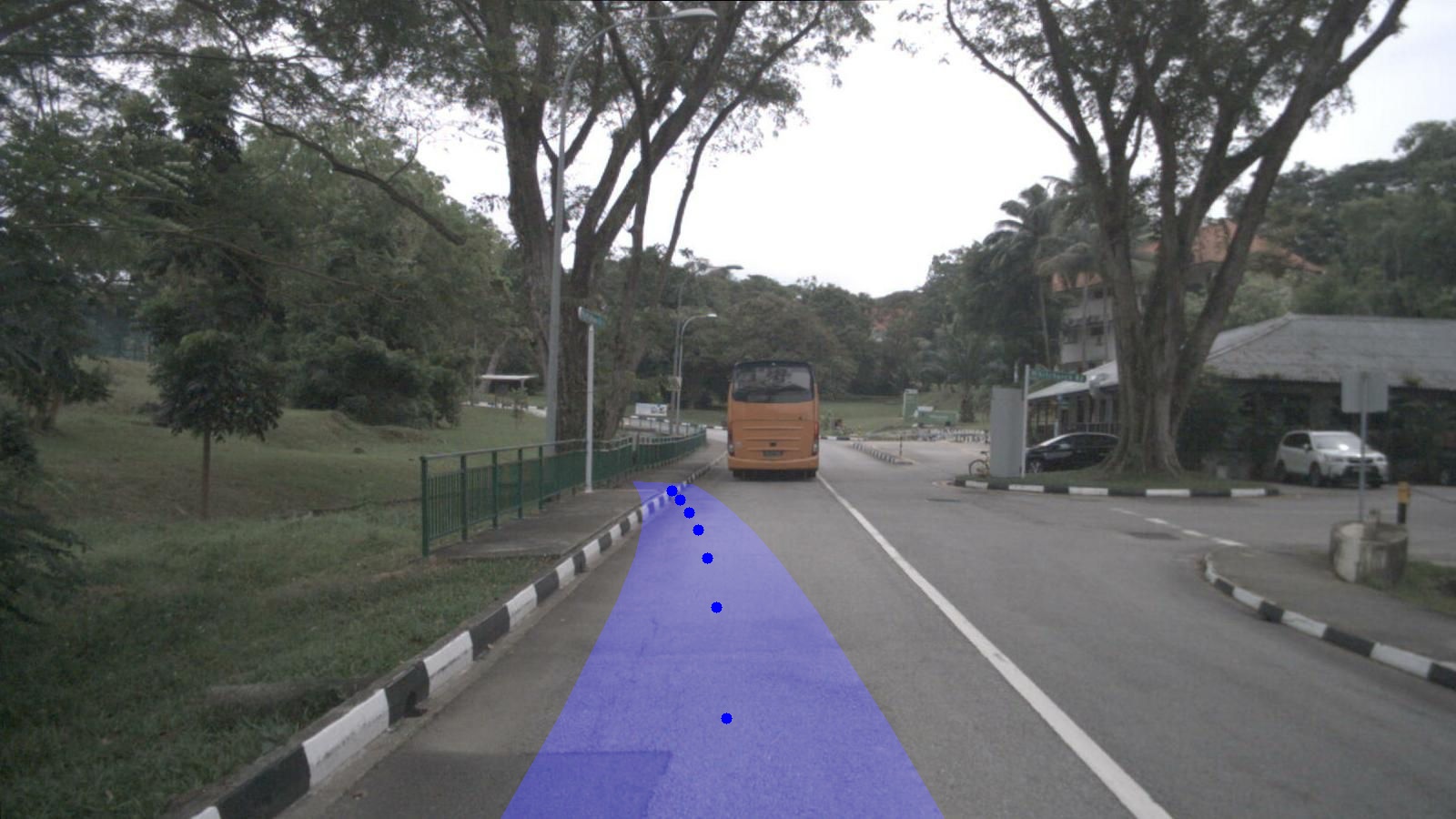}
    \begin{tcolorbox}[promptstyle, title=OpenEMMA Response (Bus Scene)]
    % \title{hal\_1}
Scene Description: As an autonomous driving labeller, I would describe the driving scene based on the provided image and the information about the camera's history. Here is the description:

1. **Traffic Lights**: There are no visible traffic lights in the image.
    2. **Movements of Other Cars or Pedestrians**: \textcolor{red}{There are no other cars or pedestrians visible in the image}.
    3. **Lane Markings**: The road has clear lane markings, including a solid white line on the right side \textcolor{red}{and a dashed white line on the left side}. There is also a black and white striped curb on the left side of the road.

The image shows a quiet
Object Description: In the driving scene, there are several road users that you should pay attention to:

1. **Bus (Location: Center of the image, moving towards the camera)**:
   - **Description**: \textcolor{red}{The bus is moving towards you on the same road.} It is important to monitor its speed and direction to ensure safe overtaking or passing.
   - **Why it's important**: Ensuring you have enough space to overtake safely is crucial to avoid collisions.
...
% 2. **Vehicles parked on the right side (Location: Right side of the image, near the curb)**:
%    - **Description**: There are vehicles parked on the right
% Intent Description: Based on the image and the information provided, the current intent to turn left is still appropriate. The ego car is in the correct lane for turning left, and there are no other cars or pedestrians in the immediate vicinity that would require a change in direction. The solid white line on the right side of the lane confirms that it is the correct lane for turning left.
% Average Displacement Error: 3.130024250386151
    \end{tcolorbox}
    \caption{OpenEMMA Testing Example (Bus Scene) on the nuScenes Dataset \cite{caesar2020nuscenes}. In the front-view image, OpenEMMA’s projected future path is overlaid as light blue. An orange bus occupies the same lane, traveling in the same direction as the ego vehicle. Solid white lane markings run along the right side, with a white-striped curb on the left. OpenEMMA erroneously plans a leftward trajectory, which would result in a collision with that curb.}
    \label{fig:vlm_hal_1}
\end{figure}

Methods that use finetuned multimodal VLMs at inference time are at risk of hallucinations, which largely stem from the VLMs' limited spatial reasoning capabilities, as discussed in the Introduction of the manuscript (Sec. 1). These models are trained almost exclusively on 2D image data, i.e., without explicit depth cues provided by modalities like LiDAR, and, as such, often misjudge object orientation and distance, producing descriptions that contradict the scene’s geometry. We show three concrete examples of VLM hullucination from OpenEMMA~\cite{xing_openemma_2024}, an open-source implementation of EMMA~\cite{hwang2024emma}. OpenEMMA uses the front image as the only input modality to the VLM and uses chain-of-thought to reason about the scenes, driver intent, and planning decisions. 

\begin{figure}[h]
    \centering
    \includegraphics[width=0.8\linewidth]{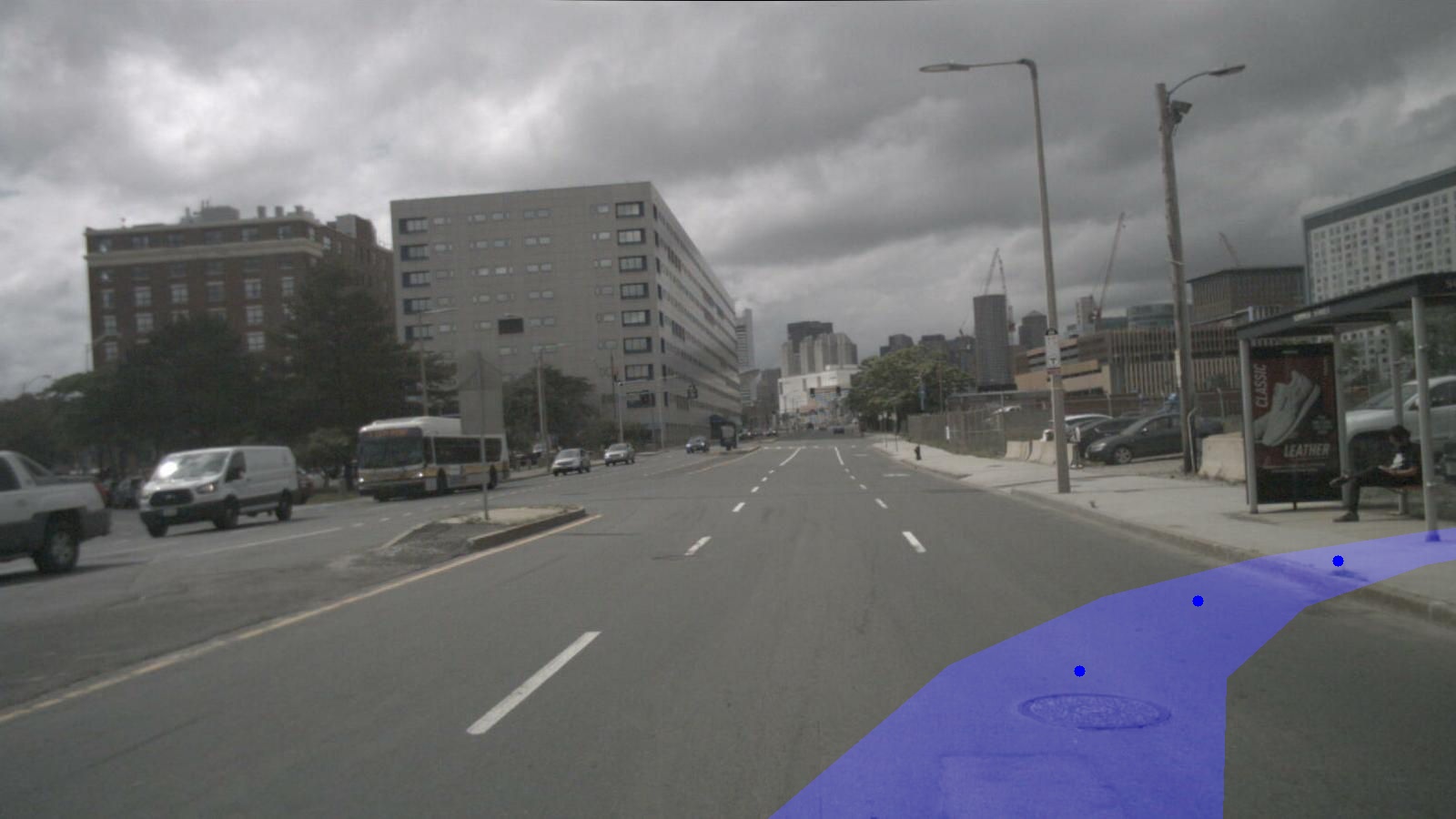}
    \begin{tcolorbox}[promptstyle, title=OpenEMMA Response(City Scene)]
Scene Description: As an autonomous driving labeller, I would describe the driving scene based on the provided images and the information available from the camera at a 0.5-second interval over the past 5 seconds.

Current Frame (0.5 seconds ago):
- **Traffic Lights:** Not visible in the current frame.
- **Other Cars:**
  - \textcolor{red}{A white van is driving on the left side of the road, moving in the same direction as the viewer}.
  - \textcolor{red}{A bus is also driving on the left side of the road, moving in the same direction as the viewer}.
  - Several cars are visible on the right side of the ...
% Object Description: In the driving scene, there are several road users that you should pay attention to:

% 1. **White Van (Location: Left Lane, Mid-Left of the Image)**:
%    - **Description**: The white van is traveling in the left lane and appears to be changing lanes. It is important to monitor its movements to ensure safe lane changes and avoid potential collisions.

% 2. **Bus (Location: Center of the Image, Mid-Left)**:
%    - **Description**: The bus is traveling in the center lane and is likely to be a public transportation vehicle. It is essential to maintain a safe distance from the bus and be prepared
% Intent Description: Based on the current information, the intent to maintain the current lane and speed is still appropriate. There are no immediate threats or changes in the lane markings that would require the ego car to change lanes or speed up. The car in front is maintaining a steady speed, and there are no pedestrians or other vehicles in the immediate vicinity that would require the ego car to change lanes or speed up. Therefore, the current intent to maintain the current lane and speed is still appropriate.
% Average Displacement Error: 9.745074706421839
\end{tcolorbox}
    \caption{OpenEMMA Testing Example (City Scene) on the nuScenes Dataset \cite{caesar2020nuscenes}. In the front-view image, OpenEMMA’s predicted trajectory is overlaid in light blue. The ego vehicle’s current lane is clear of obstacles or other traffic. On the left, a white van, a bus, and several smaller cars approach in the opposite direction in the adjacent lane. Despite this clear path, OpenEMMA incorrectly plans a rightward turn—risking a potential collision.}
        \label{fig:vlm_hal_2}

\end{figure}

\begin{figure}[t!]
    \centering
    \includegraphics[width=0.8\linewidth]{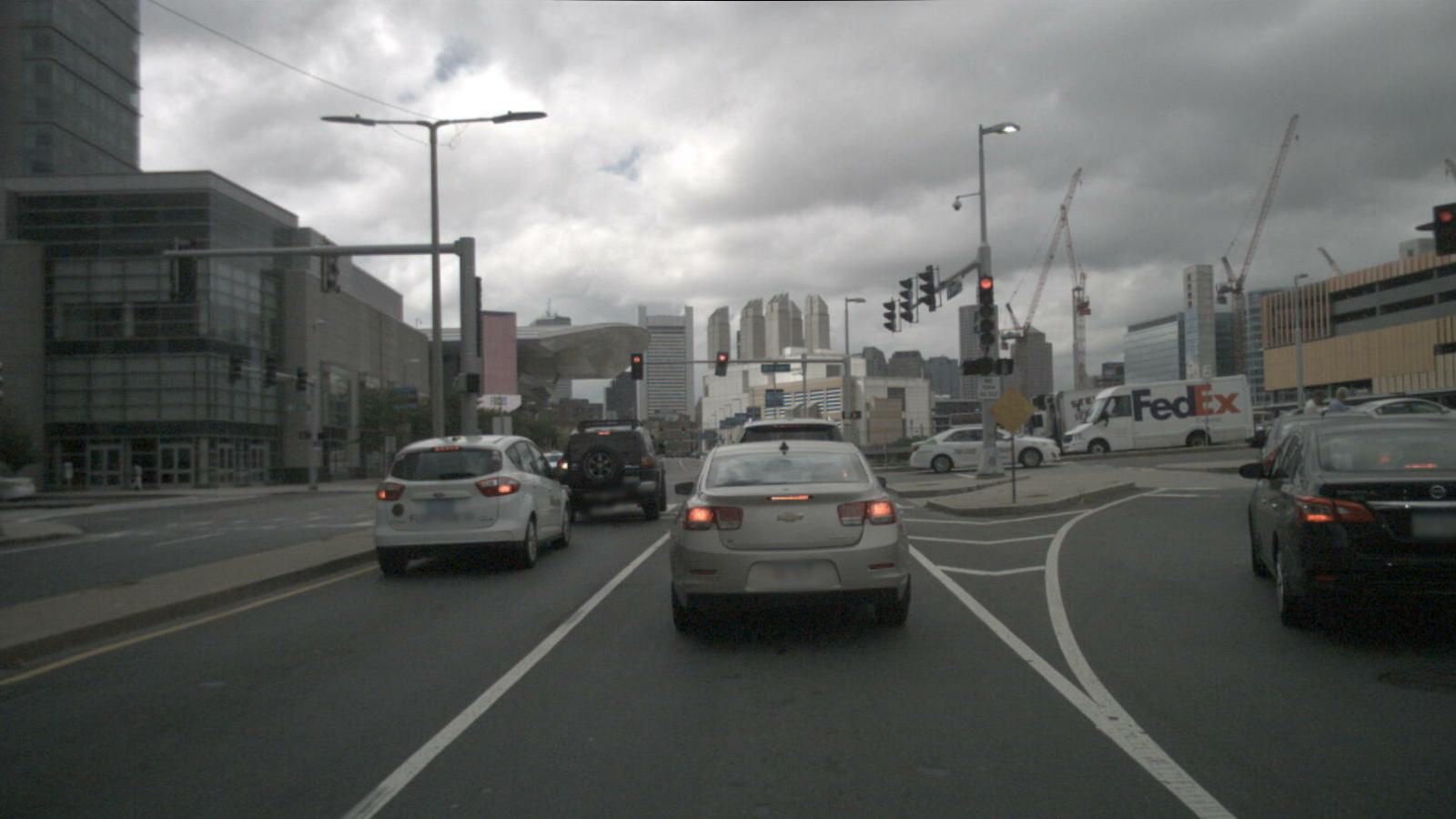}
    
\begin{tcolorbox}[promptstyle, title=OpenEMMA Response (Traffic light Scene)]
Scene Description: As an autonomous driving labeller, I would describe the driving scene based on the provided images and the information available from the camera at a 0.5-second interval over the past 5 seconds.

Current Frame (0.5 seconds ago):
- **Traffic Lights**: The traffic light in the center of the image is red, indicating that vehicles must stop.
- **Other Cars**:
  - The car directly in front of you is a silver sedan, which appears to be stationary.
  - To the left, there is a white sedan that is also stationary.
  - To the right, there is a black sedan that is
Object Description: In the driving scene, there are several road users that you should pay attention to:

1. \textcolor{red}{**FedEx Truck (Location: Right side of the image, near the center)**
   - **Description:** The FedEx truck is on the right side of the image, near the center. It is important to pay attention to this truck because it is a large vehicle that may have a longer stopping distance than smaller cars. You should be prepared for it to slow down or stop suddenly, especially if it is making a turn or changing lanes.}
...
% 2. **Construction Cranes (Location: Background, right side of the image)**
%    -
% Intent Description: Based on the updated information, I will keep my intent of turning left. Here's the reasoning:

% 1. **Lane Markings and Movement:**
%    - The ego car is in the left lane, which is the correct lane for turning left.
%    - The car in front of the ego car is slowing down and appears to be preparing to turn left, which aligns with the ego car's intention.
%    - There are no visible lane markings indicating a turn right or straight ahead, reinforcing the need to turn left.

% 2. **Other Vehicles and Pedestrians:**
%    - The car in front is slowing down, likely preparing
% Average Displacement Error: 7.42592659239553e-07
\end{tcolorbox}
    \caption{OpenEMMA Testing Example (Stop Sign Scene) on the nuScenes Dataset \cite{caesar2020nuscenes}. In the front-view image, all cars in front stop for the red traffic lights. There is no trajectory planne dand projected onto the front image. Notice, the white FeDex car on the right corner. Due to the lack of the depth and spatial understandingting, openEMMA suggested to keep an eye on this FeDex van since it thinks it could be a treat to the ego vehicle.}
    \label{fig:vlm_hal_3}
\end{figure}

In Figure~\ref{fig:vlm_hal_1}, an orange bus is driving directly in front of the ego vehicle, traveling in the same direction. However, the VLM’s response for the ``Bus Scene'' (highlighted in red) involves three clear hallucinations. It wrongly describes the bus as moving toward the viewer on the same road. Second, it invents a dashed line on the left side—there is actually only a painted curb. Third, it self-contradicts by stating that no other cars or pedestrians are visible, despite the parked cars in the upper right corner and the bus itself clearly appearing in the image. The planned intent—a slight left turn—would result in a collision with the curb and shoulder. Because the VLM  fails to recover spatial reasoning (e.g., learned in a BEV representation) needed to anticipate such obstacles, undermining both the accuracy and interpretability of using its output directly for planning.

In Figure \ref{fig:vlm_hal_2}, the ego vehicle travels alone down the road with no obstacles directly ahead. On the opposite side of the road, a white bus, a white van, and several smaller cars approach from the left in the oncoming lane. However, the VLM’s response for the ``City Scene'' wrongly reports that the bus and van are both on the left side of the road and moving in the same direction as the ego vehicle, an error that causes the planned trajectory to steer inaccurately to the right.

% In Figure \ref{fig:vlm_hal_3}, the ego vehicle travels along a clear road with no obstacles directly ahead, aside from two bicycles parked on the shoulder. It is evident that the vehicle could simply remain in the left lane and continue straight. Accordingly, under “Desired Intent,” the VLM correctly recommends staying in the left lane without changing lanes or turning. However, because the VLM relies on a single 2D image input and lacks a BEV map or any depth cues, it cannot infer the spatial boundaries of the left lane. This absence of spatial information leads to an unconstrained planning output—and ultimately a planning failure. Moreover, as marked in red, the VLM cautions against potential collisions with the bicycles despite the absence of riders, an additional hallucination that stems from its inability to distinguish between unattended objects and true hazards. 
In Figure \ref{fig:vlm_hal_3}, the ego vehicle and surrounding cars have all stopped at the red light and are waiting for it to turn green, so no trajectory is planned. However, OpenEMMA’s response highlights the white FedEx van on the right, presumably because it appears near the image center and thus seems close enough to require extra braking distance. This misattention stems from the model’s lack of explicit depth perception and spatial reasoning, which leads it to hallucinate a hazard where none exists.

\vspace{5pt}
\emph{VERDI avoids single-view hallucinations by leveraging multi-view imagery and constructing a BEV map}. From the multi-view image set, we generate a BEV representation that encodes spatial constraints, including lanes, road boundaries, other agents, and obstacles, ensuring accurate geometric context for downstream planning. Simultaneously, VERDI distills the VLM’s strong visual recognition to identify scene objects and align their textual embeddings with the projected multi-view image features, producing dense, semantically informed features that feed into the BEV encoder.

\section{Implementation Details}
\label{app:implementation_details}
\subsection{Full Prompt for VLM Queries (Section 3.2 in main).}
Below, we list the full chain-of-thought prompts used to acquire VLM reasoning embeddings. The system prompt is parsed for every prompt, and the perception, prediction, and planning prompts shown below take inputs from answers to previous modules.

\begin{tcolorbox}[promptstyle, title=System Prompt]
You are an expert driver collecting data in various scenarios for a self-driving car. Now, you are annotating the collected trajectories by providing concise and accurate descriptions as short, comma-separated sentences.
\end{tcolorbox}

\begin{tcolorbox}[promptstyle, title=Perception]
The images shown are the multi-views of a driving scenario at the beginning and end of the trajectory, 
in the order of [front left, front, front right, back left, back, back right]. 
The past \textcolor{Orchid}{\{T\}} time steps (0.5 seconds each) of the vehicle states, 
in the form of [x, y, z, vx, vy, vz], is given as: \textcolor{Orchid}{\{...\}}\\~\\
What do you see in the scene? List the following details in order:
number of lanes; agents (if any) chosen from
[car, truck, construction\_vehicle, bus, trailer, barrier, motorcycle, bicycle, pedestrian, traffic\_cone];
Location of the objects (if any) chosen from [front, left, left-front, left-behind, right, right-front, right-behind, behind];
List them from left to right and front to back.
Only list the objects that are visible in the images. Add one descriptive word for each object, e.g., red car.
\end{tcolorbox}

\begin{tcolorbox}[promptstyle, title=Prediction]
The images shown are the front views of a driving scenario at the beginning and end of the trajectory. The past \textcolor{Orchid}{\{T\}} time steps (0.5 seconds each) of the vehicle states, 
in the form of [x, y, z, vx, vy, vz], is given as: \textcolor{Orchid}{\{...\}}\\~\\
What actions did other agents take? Describe in the same order they were listed.\\~\\
Perception: \textcolor{Orchid}{\{answer to perception prompt\}}
\end{tcolorbox}

% \begin{tcolorbox}[promptstyle, title=Planning]
% The images shown are the front views of a driving scenario at the beginning and end of the trajectory. The past \textcolor{Orchid}{\{T\}} time steps (0.5 seconds each) of the vehicle states, 
% in the form of [x, y, z, vx, vy, vz], is given as: \textcolor{Orchid}{\{...\}}\\~\\
% What actions did you take? Please list all the actions taken chosen from the list: 
% \{go straight, speed up, maintain speed, move slowly, stop smoothly, stop abruptly, reverse, 
% turn left slightly, turn on blinker, turn right slightly, turn left abruptly, turn right abruptly, 
% turn around, merge into the left lane, merge into the right lane\}\\~\\
% Perception: \textcolor{Orchid}{\{answer to perception prompt\}}\\
% Prediction: \textcolor{Orchid}{\{answer to prediction prompt\}}
% \end{tcolorbox}

% \textbf{PFP Details.}
\subsection{Progressive Feature Projector (PFP) for Perception (Section 3.3 in main)}
We introduce Progressive Feature Projectors (PFP) to progressively compress and project perception, prediction and planning features of end-to-end models
After obtaining the results from implicit e2e features to language aligned feature spaces.

\begin{table}[h]
    \centering
    \begin{tabular}{c l c l}
        \multicolumn{4}{l}{\large $\text{PFP}^{\texttt{perception}}_\phi$} \\
        \hline
        & Layer & Activation & Output Dimension \\
        \hline
        0 & Input: $\mathbf{F}^{\mathtt{perception}}$ & - & $\mathbb{R}^{B\times6\times256\times12\times20}$ \\
        1 & Conv3D & ReLU & $\mathbb{R}^{B\times256\times1\times12\times20}$ \\ 
        2 & $\text{Conv2D}_{s=2,p=1}$ & ReLU & $\mathbb{R}^{B\times256\times6\times10}$ \\ 
        3 & $\text{Conv2D}_{s=2,p=1}$ & ReLU & $\mathbb{R}^{B\times256\times3\times5}$ \\ 
        4 & $\text{Conv2D}_{s=1,p=1}$ & ReLU & $\mathbb{R}^{B\times256\times3\times5}$ \\ 
        5 & Reshape & - & $\mathbb{R}^{B\times(256\cdot3\cdot5)}$ \\
        6 & Linear & - & $\mathbb{R}^{B\times D}$ \\
        \hline
    \end{tabular}
    \vspace{6pt}
    \caption{\textbf{Architecture of the PFP for Perception}. To achieve alignment between language and perception features the presented architecture compresses and extracts language space-aligned information from the perception features of the e2e model.}
    \label{tab:pfp_perception}
\end{table}

\begin{tcolorbox}[promptstyle, title=Planning]
The images shown are the front views of a driving scenario at the beginning and end of the trajectory. The past \textcolor{Orchid}{\{T\}} time steps (0.5 seconds each) of the vehicle states, 
in the form of [x, y, z, vx, vy, vz], is given as: \textcolor{Orchid}{\{...\}}\\~\\
What actions did you take? Please list all the actions taken chosen from the list: 
\{go straight, speed up, maintain speed, move slowly, stop smoothly, stop abruptly, reverse, 
turn left slightly, turn on blinker, turn right slightly, turn left abruptly, turn right abruptly, 
turn around, merge into the left lane, merge into the right lane\}\\~\\
Perception: \textcolor{Orchid}{\{answer to perception prompt\}}\\
Prediction: \textcolor{Orchid}{\{answer to prediction prompt\}}
\end{tcolorbox}

Especially in the perception stage of autonomous driving e2e models feature maps, such as intermediate features of the BEV encoder, e.g. BEVformer~\cite{li2024bevformer} in VAD~\cite{jiang2023vad}, become very large. For this stage we introduce a unique PFP to progressively compress the large feature map while preserving the feature information needed. We present an overview  of the implementation used in VERDI in Tab.~\ref{tab:pfp_perception}

%BEVformer~\cite{li2024bevformer}, which is a common architecture in autonomous driving and used by VAD, we need to squeeze the feature map for the subsequent prediction. As the prediction stage needs a two-dimensional feature sequence, we introduce a PFP to progressively compress this huge feature map while preserving the feature information needed. 

Initially a batch of multi-view image features with the shape of $\mathbf{F_i}^{\mathtt{perception}} \in \mathbb{R}^{B \times V \times C \times H \times W},$
where \(B\) is the batch size, \(V=6\) is the number of surrounding camera views, \(C=256\) is the channel dimension, and \(H=12,\ W=20\) are the spatial dimensions after the image feature extraction. The Input features are then compressed across there view dimension using a 3D Convolution.
    % \[
    %   \mathbf{F}^{(1)} = \mathrm{ReLU}\bigl(\mathrm{Conv3D}\bigl(\left( \mathbf{F_i}^{\mathtt{perception}} \right)\bigr)\bigr)
    %   \in \mathbb{R}^{B\times256\times1\times12\times20}.
    % \]
Then the fused map is processed by a sequence of 2D convolutional layers to reduce the spatial dimension and extract essential information. A final 2D convolution refines these coarse spatial features without further down-sampling. %as a post-processing for this sub-stage:
    % \[
    %   \mathbf{F}^{(3)} = \mathrm{ReLU}\bigl(\mathrm{Conv2D}_{s=2,p=1}(\mathbf{F}^{(2)})\bigr)
    %   \;\in\;\mathbb{R}^{B\times256\times6\times10},
    % \]
    % \[
    %   \mathbf{F}^{(4)} = \mathrm{ReLU}\bigl(\mathrm{Conv2D}_{s=2,p=1}(\mathbf{F}^{(3)})\bigr)
    %   \;\in\;\mathbb{R}^{B\times256\times3\times5}, 
    % \],
    % \[
    %   \mathbf{F}^{(5)} = \mathrm{ReLU}\bigl(\mathrm{Conv2D}_{s=1,p=1}(\mathbf{F}^{(4)})\bigr)
    %   \in \mathbb{R}^{B\times256\times3\times5}.
    % \]
Finally, we reshape the feature map into a flat vector for the subsequential processing
    % \[
    %   \mathbf{f} = \mathrm{vec}\bigl(\mathbf{F}^{(5)}\bigr) \in \mathbb{R}^{B\times(256\cdot3\cdot5)},
    % \]
    and apply a linear projection to obtain the final compressed feature. 
    % \[
    %   f^{\texttt{perception}}_{\mathcal P} = \mathbf{W}\,\mathbf{f} + \mathbf{b}
    %   \;\in\;\mathbb{R}^{B\times D}.
    % \]
This feature will be used to align with the encoded feature map from VLM's annotation, where $D$ represents the size of the shared feature space.

\subsection{Fully Connected Layers (MLP) for Prediction and Planning (Section 3.3 in main)}
In contrast to perception, the prediction and planning features of the e2e model do not have a natural two- or three-dimensional spatial structure. These stages feature sequences are of the form $ \mathbf{F_i}\in\mathbb{R}^{B\times S}$. We therefore compress respective features into a representation of dimension $D$ that can be aligned with VLM‐processed features adopting a stack of $L$ fully‐connected (MLP) layers, each followed by Layer Normalization and a ReLU activation.

All linear projection layers are are implemented with normalization and a ReLU nonlinearity as described in Tab.~\ref{tab:pfp_pred_plan}. Ultimately the output is again projected to the hidden dimension $D_{\mathrm{hidden}}$. %:
% \[
% \mathbf{F}^{(1)} = \mathrm{ReLU}\Bigl(\mathrm{LayerNorm}\bigl(\mathbf{W}^{(1)}\, \mathbf{F_i} + \mathbf{b}^{(1)}\bigr)\Bigr)
% \;\in\;\mathbb{R}^{B\times D_{\mathrm{hidden}}}.
% \]
% Then, for each layer $l=2,\dots,L$, the feature is refined as:

% \[
% \mathbf{F}^{(l)} = \mathrm{ReLU}\Bigl(\mathrm{LayerNorm}\bigl(\mathbf{W}^{(l)}\,\mathbf{F}^{(l-1)} + \mathbf{b}^{(l)}\bigr)\Bigr)
% \;\in\;\mathbb{R}^{B\times D_{\mathrm{hidden}}},\quad l=2,\dots,L.
% \]
% Finally, a linear “head” maps the last hidden feature to the target dimension $D$:

% \[
% f^{\texttt{}}_{\mathcal P} = \mathbf{W}^{(o)}\,\mathbf{F}^{(L)} + \mathbf{b}^{(o)}
% \;\in\;\mathbb{R}^{B\times D}.
% \]

This output $f^{\texttt{}}_{\mathcal P}$ serves as the unified feature representation for the downstream prediction and planning module. Note that for prediction and planning stage we will use two separate models, that do not share any parameters as they are handling different tasks respectively.

\begin{table*}[h]
    \centering
    \begin{tabular}{c l c c l}
        \multicolumn{5}{l}{\large $\text{PFP}^{ \{ \texttt{prediction}\text{, } \texttt{planning}\}}_\phi$} \\
        \hline
        & Layer & Normalization & Activation & Output Dimension \\
        \hline
        0 & Input: $\mathbf{F}^{\mathtt{i}}$ & - & - & $\mathbb{R}^{B\times6\times S}$ \\
        1 & Linear & LayerNorm & ReLU & $\mathbb{R}^{B\times D_{\mathrm{hidden}}}$ \\ \\
        l = 2, ... L-1 & \multicolumn{4}{c}{ ... } \\ \\
        L & Linear & LayerNorm & ReLU & $\mathbb{R}^{B\times D_{\mathrm{hidden}}}$ \\ 
        L+1 & Linear & - & - & $\mathbb{R}^{B\times D}$ \\
        \hline
    \end{tabular}
    \vspace{6pt}
    \caption{\textbf{Architecture of the PFP for Prediction and Planning}. To achieve alignment between language and prediction or planning features the presented architecture compresses and extracts language space-aligned information from the respective e2e modules.}
    \label{tab:pfp_pred_plan}
\end{table*}

\section{Ablation Details and VLM Embedding Quality (Section 4.3 in main)}
\label{app:ablation_loss_and_details}
\subsection{Evaluation Setting}
\begin{figure}[h]
    \centering
    \includegraphics[width=0.9\linewidth]{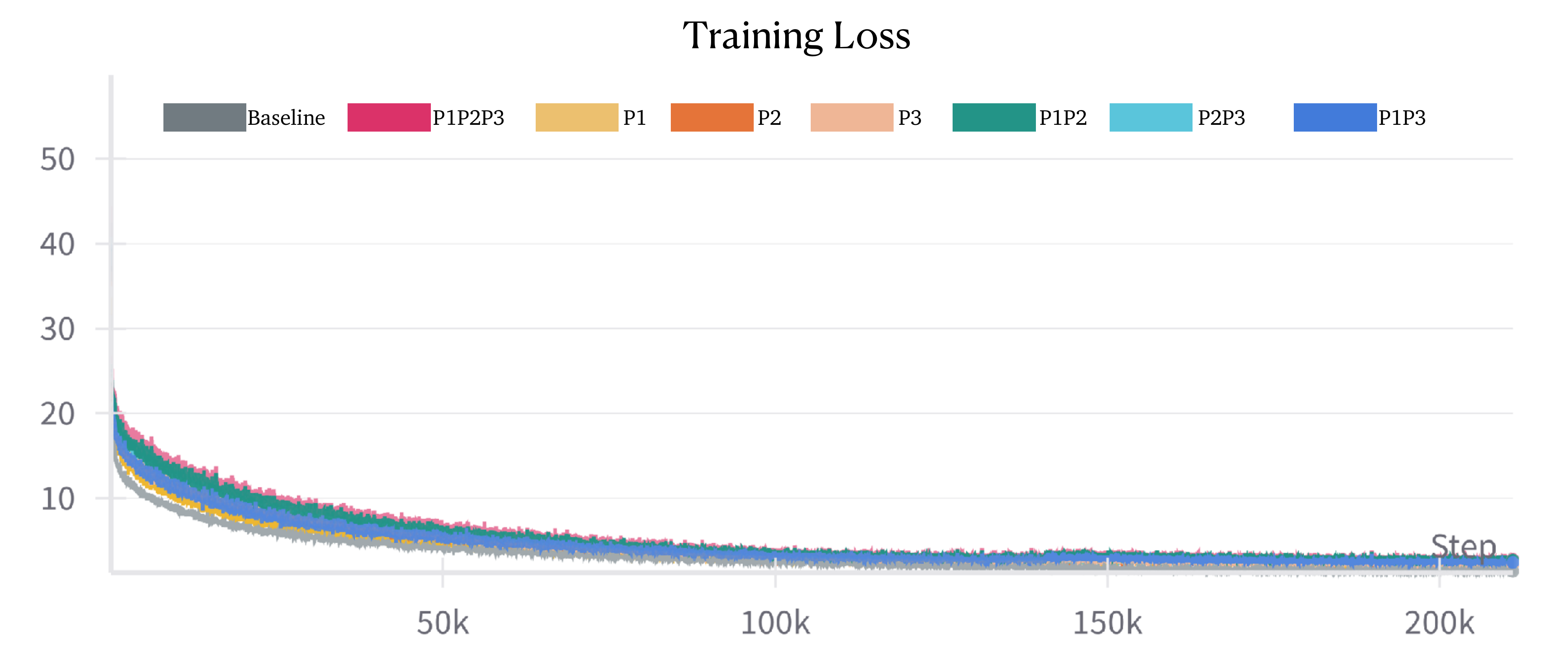}
    \caption{Training loss over gradient steps for all module variations (each trained for 30 epochs). We observe that as more modules are aligned, the training loss increases. Here, P1, P2, and P3 refer to alignment of the perception, prediction, and planning modules, respectively. The ablation study evaluates different combinations of these alignments, and in every case the loss curves plateau on this small ablation dataset, indicating training saturation.}
    \label{fig:module_sat}
\end{figure}

We create a smaller dataset for the ablation experiments that require retraining all ablated variants. For the module variations ablation study, we aim to reduce training time while preserving the full dataset’s diversity. This ablation dataset contains trajectories from each scene in the full dataset that begin at $t$ = $0$, $10$, and $20$ and run for $10$ time steps. Since most scenes span around $40$ time steps, these samples evenly cover the entire dataset with no overlaps.  P1, P2 and P3 (see also Fig~\ref{fig:module_sat}) means perception, prediction and planing alignment separately. All models are trained for $30$ epochs and evaluated on the full test set. We show that both VERDI and baseline model trainings saturate on this ablation dataset in Fig~\ref{fig:module_sat}, making the comparison fair. VERDI shows $8.2\%$ improvement over the baseline method (VAD-tiny), confirming the effectiveness of distilling VLM's reasoning capability to the driving modules.
% Additional qualitative comparisons between VERDI and the supervised e2e baseline.
% \begin{figure}[b!]
%     \centering
%     \includegraphics[width=\linewidth]{figures/additional_viz.pdf}
%     \caption{Additional qualitative comparison of \textsc{VERDI} (Ours, right column) and the Supervised e2e model (baseline, left column) on the nuScenes dataset \cite{caesar2020nuscenes}. Each entry shows the multi-view camera observations on the left and the BEV view on the right at one time step $t$. The left panel overlays the ego agent’s planned 3-second trajectory on the front-camera image and BEV panel as a solid green line that fades to blue. The BEV panel renders the ego vehicle as a green rectangle, pedestrians and other vehicles as red rectangles, and their predicted 3-second trajectories as red lines. %The ego agent’s own planned trajectory is projected onto the BEV as a green-to-blue fading line.
%     Each of the example shows our successful performance on the perception, prediction, and planning modules, indicated by \textcolor{green}{$\square$}, while failures are highlighted by \textcolor{red}{$\square$}. We also show the VLM texts response for each testing case. We observed that the VERDI's behavior highly aligns with the VLM's response, showing the successful VLM reasoning and commonsense distillation.}
%     \label{fig:additional-viz}
% \end{figure}

\subsection{VLM Embedding Quality}
Next, we evaluate the VLM embedding quality, and, following our module ablation setup, we select three trajectories of ten time steps each. To evaluate the VLM’s reasoning consistency, we start those trajectories at $t = 0$, $5$, and $10$, which creates five overlapping time steps between each pair. The intuition behind these time step choices is to amplify the impact of embedding quality: a large VLM with strong reasoning capabilities should yield consistent outputs across overlapping time steps—thereby improving performance—whereas a smaller VLM with weaker reasoning will likely produce inconsistent results on the same time steps, degrading performance. The findings confirm that with the Qwen $75$B model and the base sentence-transformer encoder, we achieve the largest improvement $6.7\%$ compared to the baseline (VAD-tiny).

\section{Additional Qualitative Results on NuScenes}
\begin{figure*}[t]
    \centering
    \includegraphics[width=\linewidth]{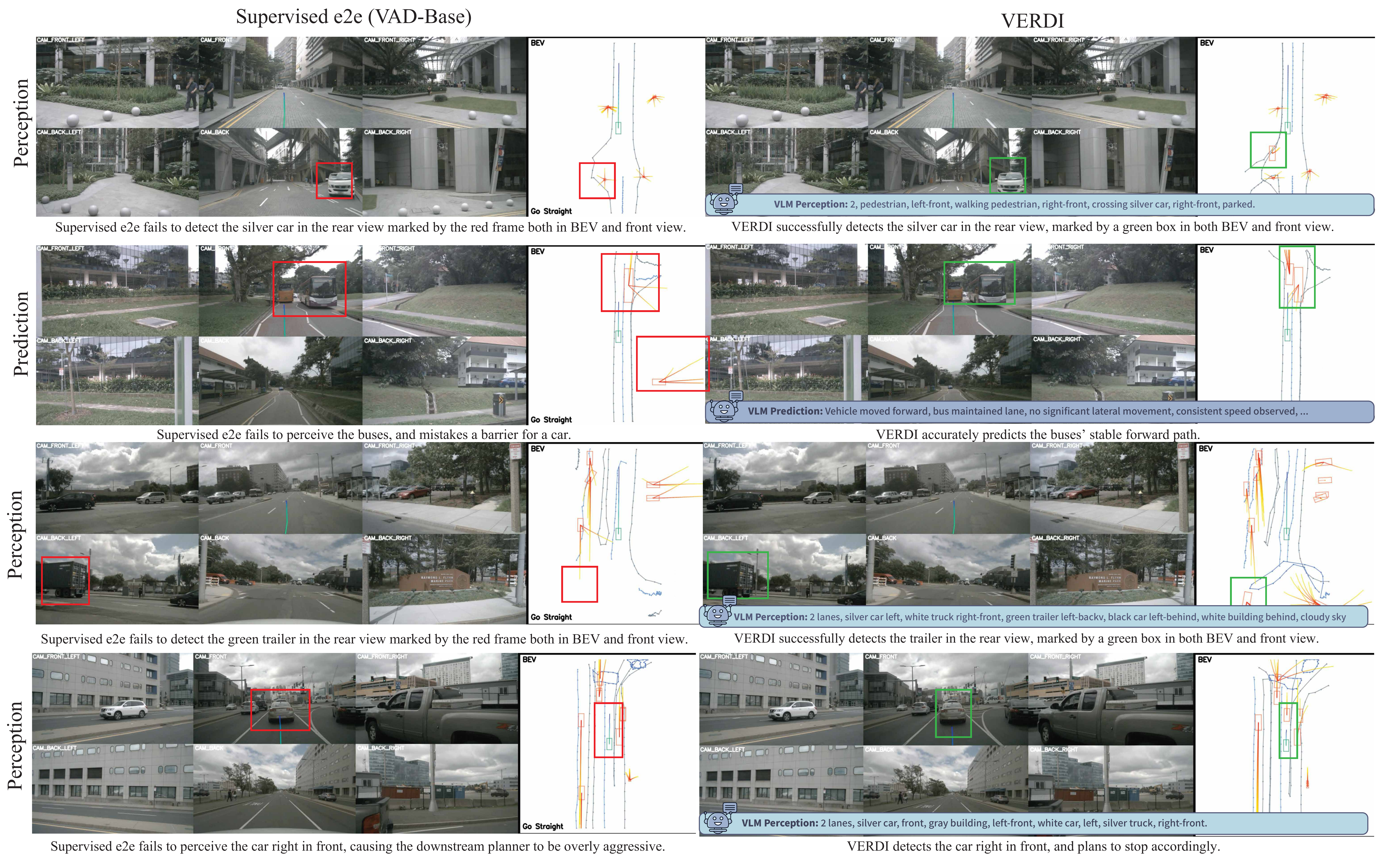}
    \caption{Additional qualitative comparison of \textsc{VERDI} (Ours, right column) and the Supervised e2e model (baseline, left column) on the nuScenes dataset \cite{caesar2020nuscenes}. Each entry shows the multi-view camera observations on the left and the BEV view on the right at one time step $t$. The left panel overlays the ego agent’s planned 3-second trajectory on the front-camera image and BEV panel as a solid green line that fades to blue. The BEV panel renders the ego vehicle as a green rectangle, pedestrians and other vehicles as red rectangles, and their predicted 3-second trajectories as red lines. Each example shows successful performance on perception, prediction, and planning modules, indicated by \textcolor{green}{$\square$}, while failures are highlighted by \textcolor{red}{$\square$}. We also show the VLM textual response for each testing case, where VERDI’s behavior closely aligns with the VLM’s reasoning and commonsense analysis.}
    \label{fig:additional-viz}
\end{figure*}
We present additional visualization results in Figure~\ref{fig:additional-viz}, illustrating four scenarios in which the supervised e2e model fails—either by missing agents on the road, mispredicting their future trajectories, or planning a path that leads to a collision due to perceptual errors. In contrast, VERDI correctly detects the vehicles behind in the first and third rows (marked by \textcolor{green}{$\square$}) and accurately predicts the forward motion of the yellow and white buses (second row, \textcolor{green}{$\square$}). Most interestingly, in the fourth row, VERDI comes to a reasonable stop at the red traffic light because it perceives the car ahead, whereas the baseline model overlooks this vehicle and proceeds forward, resulting in a collision (marked by \textcolor{red}{$\square$}). We also queried the VLM on these cases and observed that VERDI’s behavior aligns closely with the VLM’s response, further confirming successful VLM reasoning distillation.
% \begin{figure}[b]
%     \centering
%     \includegraphics[width=\linewidth]{figures/additional_viz.pdf}
%     \caption{Additional qualitative comparisons between VERDI and the supervised e2e baseline.}
%     \label{fig:additional-viz}
% \end{figure}

\section{Additional Closed-Loop HugSim Results}
\label{app:hugsim_qualitative}

\begin{figure}[H]
    \centering
    \includegraphics[width=0.8\linewidth]{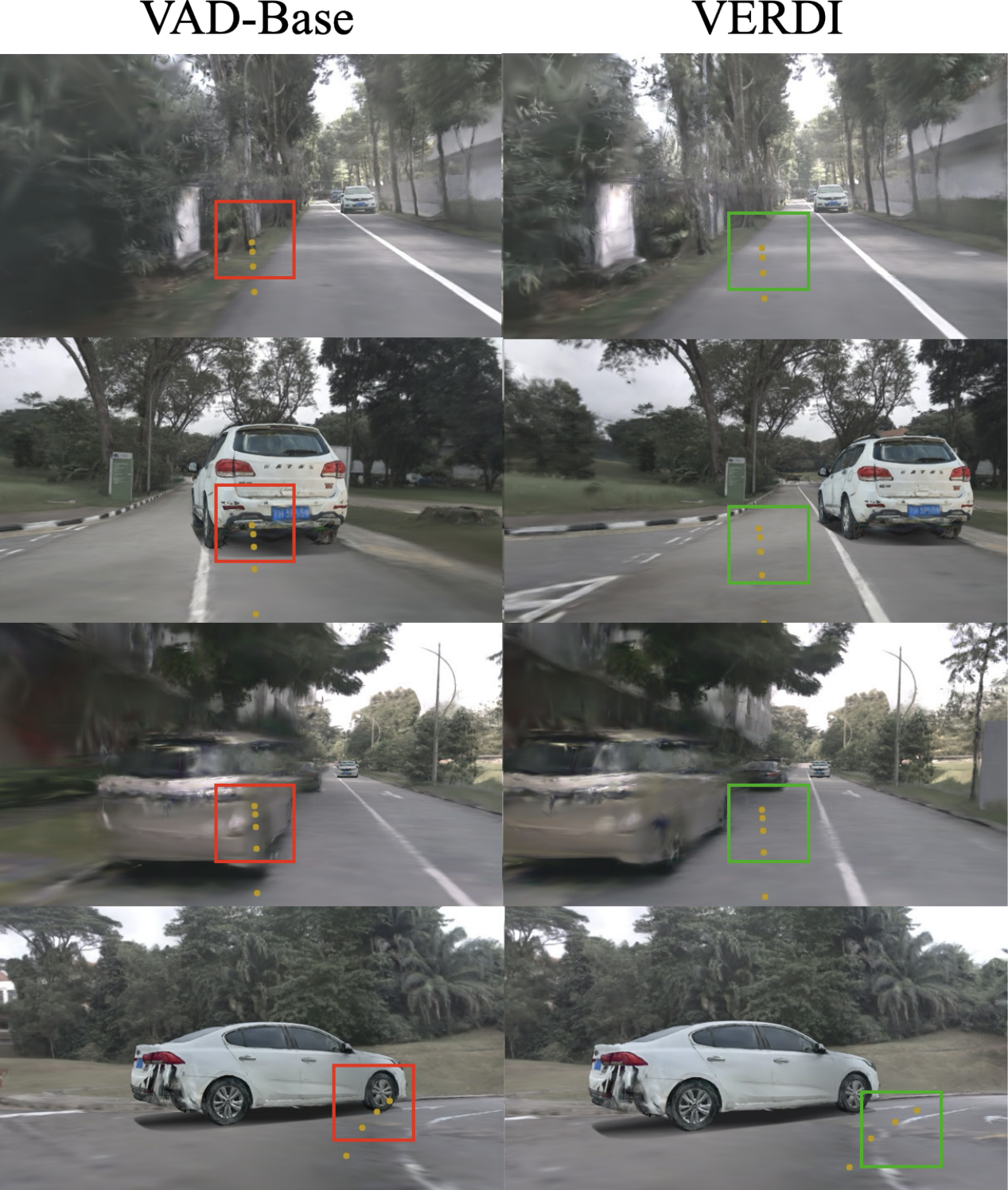}
    \caption{Qualitative comparison of \textsc{VERDI} (Ours, right column) and the Supervised e2e model (baseline, left column) in the HugSim simulator~\cite{zhou2024hugsim} close loop test. Each entry shows the front view camera observations at one time step $t$. The left panel overlays the ego agent’s planned 3-second trajectory on the front-camera image as sequential yellow dots as waypoints. Each example shows successful performance on end to end planning, indicated by \textcolor{green}{$\square$}, while failures are highlighted by \textcolor{red}{$\square$}.}
    \label{fig:hugsim-viz}
\end{figure}

As VERDI achieves a higher Non-Collision (NC) score in the HugSim closed-loop evaluation, we further analyze which types of scenarios benefit most from its improved collision-avoidance behavior. Figure~\ref{fig:hugsim-viz} presents qualitative visualizations of four challenging cases in which the VAD-Base (supervised end-to-end model) fails to plan a collision-free trajectory. In contrast, VERDI—equipped with distilled VLM reasoning, correctly detects nearby vehicles and accurately predicts their motion and plans a collision-free trajectory.

% \begin{figure*}[t]
%     \centering
%     \begin{minipage}{0.7\linewidth} % << match your image width
%         \centering
%         \includegraphics[width=\linewidth]{figures/hugsim_qal.pdf}
%         \caption{Qualitative comparison of \textsc{VERDI} (Ours, right column) and the Supervised e2e model (baseline, left column) in the HugSim simulator~\cite{zhou2024hugsim} close-loop test. Each entry shows the front-view camera observations at time step $t$. The left panel overlays the ego agent’s planned 3-second trajectory on the image as sequential yellow waypoint dots. Successful performances are indicated by \textcolor{green}{$\square$}, while failures are highlighted by \textcolor{red}{$\square$}.}
%         \label{fig:hugsim-viz}
%     \end{minipage}
% \end{figure*}

These examples represent challenging situations: one lane-following case with an oncoming vehicle approaching closely, and three bypassing cases with vehicles in tight proximity to the ego agent. The turning-while-bypassing scenario in the fourth row is particularly interesting. Such aggressive driving behavior rarely appears in the nuScenes dataset; as a result, the purely behavior-cloned VAD-Base, which lacks distilled reasoning capability, has no prior experience to guide its decision making. When encountering this scenario in the closed-loop simulator—one that exceeds the difficulty of its training set—the VAD-Base fails to produce a collision-free plan because it cannot actively reason about the scene. In contrast, VERDI handles this difficult corner case successfully, further confirming that its distilled reasoning capability enables a safer behavior in scenarios beyond the training set.